\documentclass[letterpaper]{article} 
\usepackage{aaai23}  
\usepackage{times}  
\usepackage{helvet}  
\usepackage{courier}  
\usepackage[hyphens]{url}  
\usepackage{graphicx} 
\usepackage{natbib}  
\usepackage{caption} 
\usepackage{algorithm}
\usepackage{algorithmic}

\newcommand{\eg}{{\em e.g.}}
\newcommand{\ie}{{\em i.e.}}

\usepackage{newfloat}
\usepackage{listings}

\DeclareCaptionStyle{ruled}{labelfont=normalfont,labelsep=colon,strut=off} 
\floatstyle{ruled}
\newfloat{listing}{tb}{lst}{}
\floatname{listing}{Listing}
\title{Adaptive Mixing of Auxiliary Losses in Supervised Learning}
\author{
        Durga Sivasubramanian\textsuperscript{\rm 1, \rm 2}\thanks{Work partially done while at Google Research}, 
        Ayush Maheshwari\textsuperscript{\rm 1},
        Prathosh AP\textsuperscript{\rm 2},\\
        Pradeep Shenoy\textsuperscript{\rm 3}, 
        Ganesh Ramakrishnan\textsuperscript{\rm 1}
        \\
}
\affiliations{
    \textsuperscript{\rm 1}Indian Institute of Technology Bombay \hspace{0.1cm} 
    \textsuperscript{\rm 2}Google Research, India  \hspace{0.1cm}
    \textsuperscript{3}Indian Institute of Science, Bengaluru\\
     \{durgas, ayusham, ganesh\}@cse.iitb.ac.in,  prathosh@iisc.ac.in, 
     shenoypradeep@google.com
}

\newcommand{\model}{\mbox{\textsc{Amal}}}

\usepackage{multirow}
\usepackage{xspace}

\newcommand{\bfl}{\mathbf l}
\newcommand{\bfx}{\mathbf x}

\usepackage{amsmath}
\usepackage{amssymb}
\usepackage{mathtools}
\usepackage{amsthm}
\usepackage{subcaption}
\usepackage{color}
\usepackage{subdef}
\usepackage{booktabs}       
\usepackage{adjustbox}

\newtheorem{theorem}{Theorem}
\newtheorem{lemma}{Lemma}

\usepackage{bibentry}

\begin{document}

\maketitle

\begin{abstract}
In several supervised learning scenarios, \textit{auxiliary losses} are used in order to introduce additional information or constraints into the supervised learning objective. For instance, knowledge distillation aims to mimic outputs of a powerful \textit{teacher model}; similarly, in rule-based approaches, weak labeling information is provided by \textit{labeling functions} which may be noisy rule-based approximations to true labels. We tackle the problem of learning  to combine these losses in a principled manner. Our proposal, \model{}, uses a bi-level optimization criterion on validation data to learn optimal mixing weights, at an instance-level, over the training data. We describe a meta-learning approach towards solving this bi-level objective and show how it can be applied to different scenarios in supervised learning. Experiments in a number of knowledge distillation and rule denoising domains show that \model{} provides noticeable gains over competitive baselines in those domains. We empirically analyze our method and share insights into the mechanisms through which it provides performance gains.  The code for \model{} is at: \url{https://github.com/durgas16/AMAL}.
\end{abstract}

\section{Introduction}

Deep learning techniques have shown significant impact in a wide range of machine learning applications, driven primarily by the availability of large amounts of reliable labeled data~\cite{sun2017revisiting}. Despite this progress, supervised learning faces certain challenges: first, the time and effort needed to obtain large, reliable labeled datasets, and second, the limited information contained in human-annotated labels. Several approaches aim to improve generalization and sample efficiency of supervised learning by incorporating additional sources of information, or learning constraints, into the supervised learning paradigm. For instance, 
 rule-denoising techniques~\cite{NIPS2016_6709e8d6} use simple, approximate labeling rules (labeling functions) that provide weak supervision and reduce dependence on data annotation. Other work has combined learning from labeling functions with supervised learning from limited human-annotated  data~\cite{maheshwari-etal-2021-semi}--these approaches leverage the supervised learning objective to offset the noisy labels from labeling functions.  A challenge here is how to optimally combine these complementary objectives. 

\noindent Equally, cardinal labels do not capture the richness of information contained in the input data--\eg, object category labels for images of natural scenes. Some of this imprecision can be mitigated by using more nuanced `soft labels', or distributions over labels, as the target for supervision instead of the cardinal labels. Knowledge distillation (KD) \cite{hinton2015distilling}) proposes using the inherent uncertainty of a supervised model trained on cardinal labels (the `teacher model') to generate these soft labels for training, in combination with the conventional supervision loss. Indeed, recent work~\cite{pmlr-v139-menon21a} formalizes this process from a Bayesian perspective, showing that when one-hot labels are an imperfect representation of the true probability distribution, KD reduces the variance associated with probability estimates in a student model. Other work examines, from an empirical perspective, when and how distillation may improve upon training from scratch on the labels alone. For instance, an overtrained teacher will likely achieve low/zero error rates {\em w.r.t.} the (incomplete) label loss simply by overfitting on random noise in the dataset; in these circumstances, the probabilities output by the teacher do not accurately represent the underlying uncertainty, and students may be led astray.


\noindent We propose \model, an adaptive loss mixing technique for addressing the challenge of optimally combining supervised learning objectives with these varied auxiliary objectives. Our proposal is driven by the following key insight: the mixing of primary and auxiliary objectives greatly benefits by being regulated on a \textit{sample-by-sample basis}. This draws from substantial literature showing the promise of instance-reweighting, for example in handling noisy labels or outliers~\cite{castells2020superloss,l2r}. We therefore propose to learn instance-specific mixing parameters that combine complementary learning objectives. We devise a meta-learning algorithm, based on a separate \textit{validation metric}, to estimate these instance-specific parameters in an unbiased manner. We demonstrate how our method yields more accurate models when rule-based losses are mixed with limited supervision losses~\cite{maheshwari-etal-2021-semi} as well as in a 
knowledge distillation setting (KD)~\cite{hinton2015distilling}.

\begin{figure}[t]
\centering
\includegraphics[width = 0.9\linewidth, height=1cm, ]{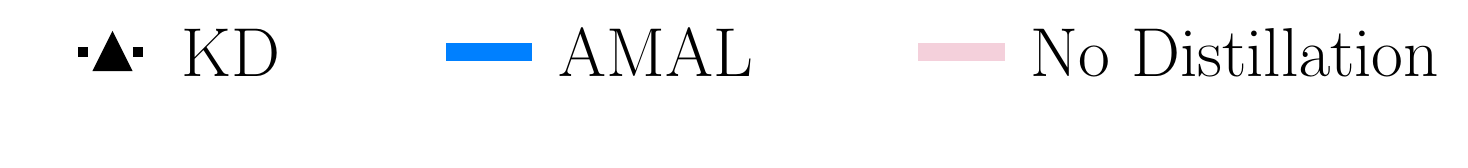}
\centering
\hspace{-0.6cm}
\begin{subfigure}[b]{0.48\linewidth}
\centering
\includegraphics[width=\textwidth, height=4cm]{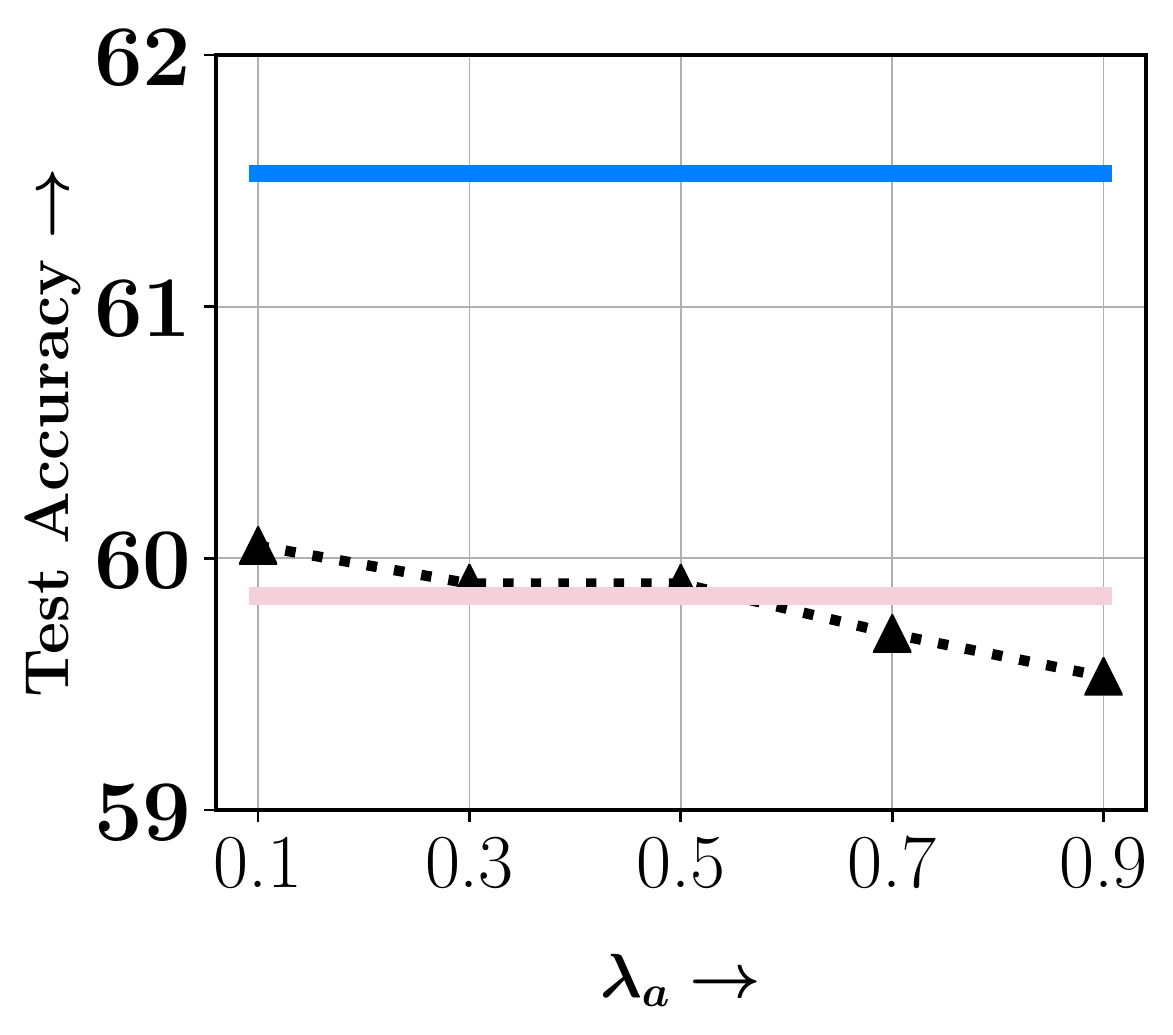}
\caption*{$\underbracket[1pt][1.0mm]{\hspace{3.5cm}}_{\substack{\vspace{-5.0mm}\\
\colorbox{white}{(a) \scriptsize Teacher-ResNet110}}}$}
\end{subfigure}
\begin{subfigure}[b]{0.48\linewidth}
\centering
\includegraphics[width=\textwidth, height=4cm]{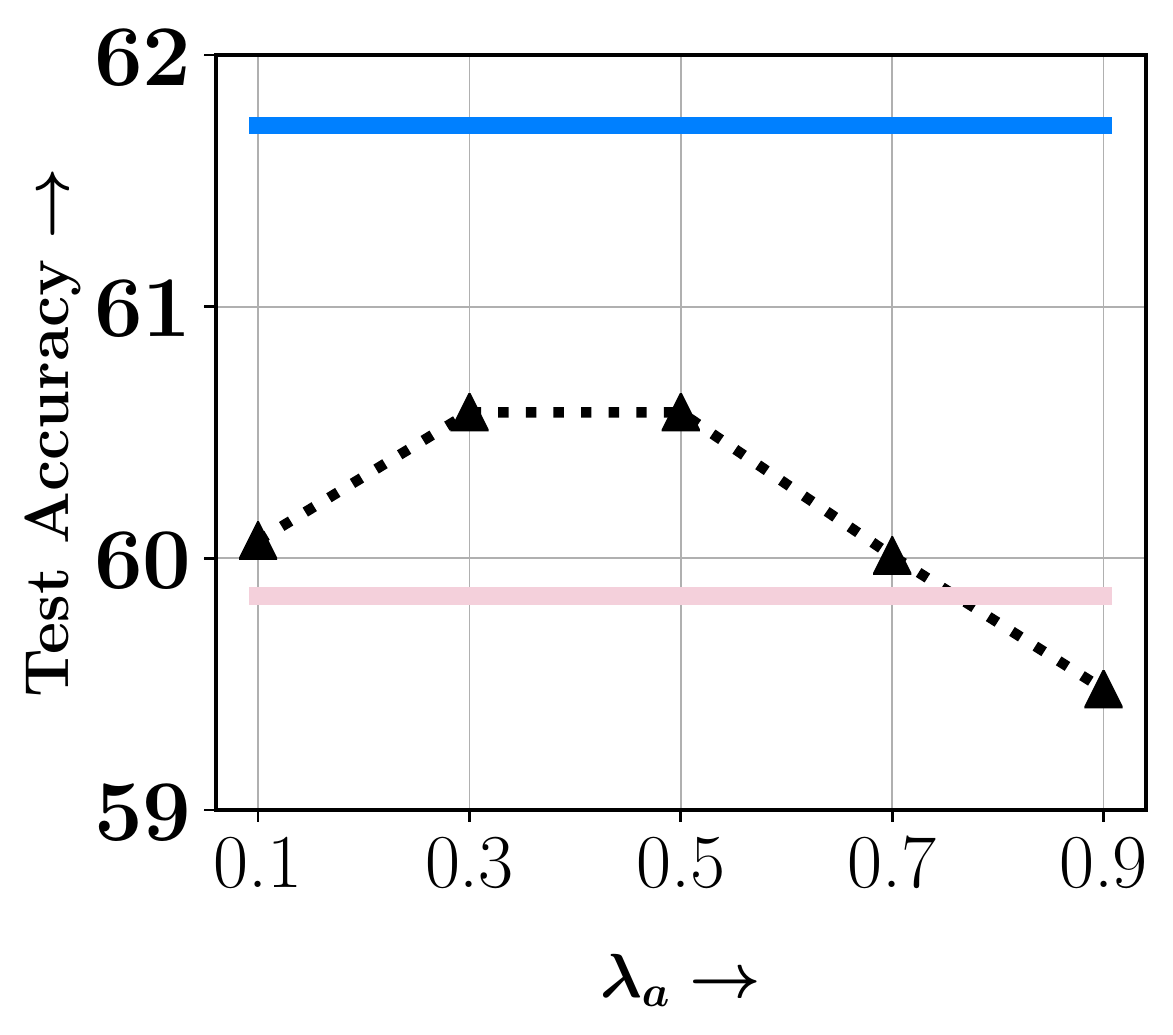}
 \caption*{$\underbracket[1pt][1.0mm]{\hspace{3.5cm}}_{\substack{\vspace{-4.0mm}\\ \colorbox{white}{(b) \scriptsize Teacher-ResNet20}}}$}
\end{subfigure}

\caption{ Knowledge distillation (KD) performed on CIFAR100, with ResNet 8 as a student model. Subfigure (a) uses ResNet110 as teacher whereas subfigure (b) uses  ResNet20 as teacher. KD performed with uniformly weighted ($\lambda_a$) performs poorly as the gap between the learning capacities of the teacher and student  models  increases. In both the cases, \model{} with the weights learned performs the best.}
\label{fig:motivatio_kd}
\end{figure}

\begin{figure}[t]
\centering
\includegraphics[width =\linewidth ]{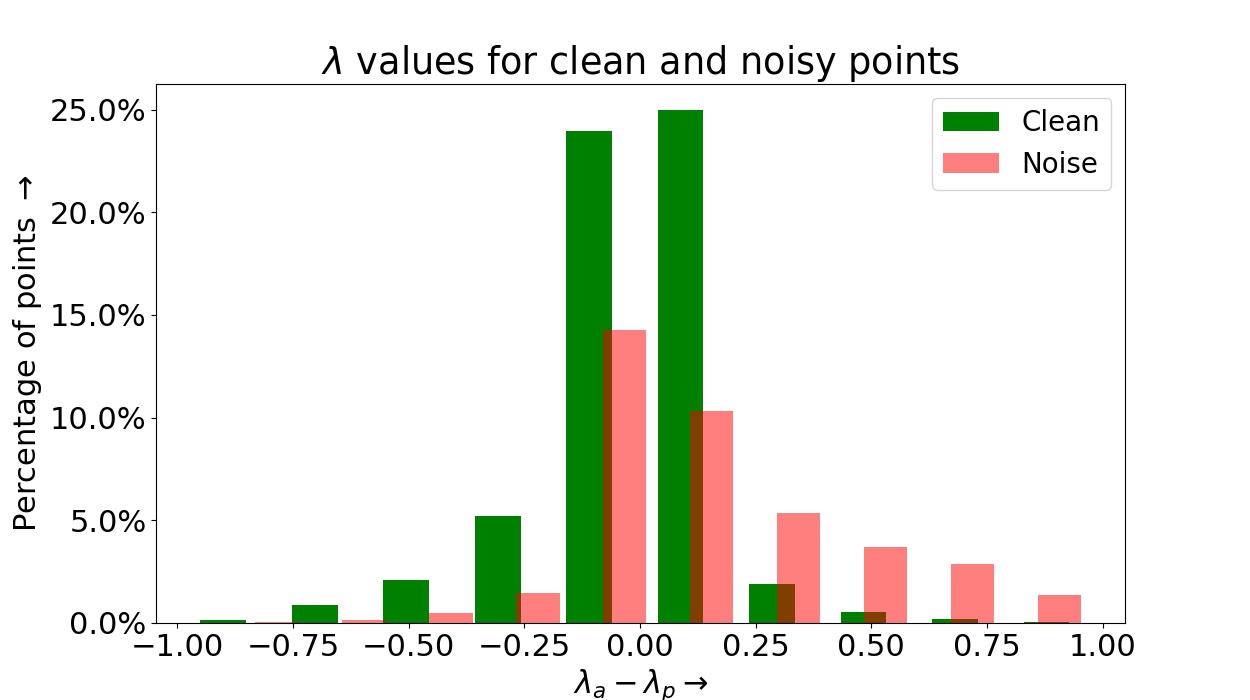}
\caption{Distribution of difference between $\lambda_a$ (weight associated with the KD loss)  and $\lambda_p$ (weight associated with the CE loss), obtained using \model{} while performing knowledge distillation with Resnet8 as the student model and Resnet110 the teacher model on CIFAR100 dataset with 40\% label noise. }
\label{distill_diff}
\end{figure}

\noindent \textbf{Motivation for our work:} We present motivation for our work in a knowledge distillation (KD) setup  on the standard CIFAR100 dataset~\cite{Krizhevsky09learningmultiple}-- the student model is set to ResNet8 and the teacher model to ResNet110 in subfigure (a). Since the capacity difference between the student and teacher models is large, mimicking the teachers outputs maybe harmful rather than helpful for the training of the ResNet8 model~\cite{cho2019efficacy}. This is illustrated in Figure~\ref{fig:motivatio_kd} where we present the performance of KD obtained with different values of $\lambda_a$ (parameter controlling the influence of KD loss - {\em c.f.} Section~\ref{sec:distillation}). Here we set $\lambda_p  = 1- \lambda_a$ as in equation \eqref{eq:distil_std}. We compare this against learning with only standard hard labels (no  KD, \ie, $\lambda_a =0$) and \model{} with learned $\lambda$s. In Figure~\ref{fig:motivatio_kd} (a) we observe that KD performs almost similar to or worse than  No-Distillation 
baseline. To understand the effect of capacity difference between the teacher and student models, in Figure~\ref{fig:motivatio_kd} (b) we perform distillation with ResNet20 as teacher and ResNet8 as student. Due to this reduction in the capacity difference, for some values of $\lambda,$ KD performs better than No-Distillation, but cannot bridge the gap to \model's performance using optimal loss mixing.

To further motivate instance-wise mixing, we apply \model{} to a KD setup with 40\% label noise injected into the CIFAR100 dataset. 
Here, too, we use ResNet110 as teacher and ResNet8 as student. We examine the difference between the weights associated with the distillation loss and the supervision loss ($\lambda_a$ and $\lambda_p$ respectively). Figure \ref{distill_diff} shows this difference as a histogram over instances separated into clean and noisy labels. \model\ favors supervision loss for clean data points (\ie, negative range of $\lambda_a - \lambda_p$ ), as intended from an optimal mixing perspective. This is consistent with the observation we made on Figure~\ref{fig:motivatio_kd}, where the student learns better when the KD loss is assigned lower weightage.
 In a similar fashion, \model\ emphasizes KD loss for noisy points, correctly identifying that the teacher model is more informative for those points than their misleading hard labels. 

\noindent \textbf{Our Contributions:} Our key contributions are as follows:
    \noindent 1) We propose a general formulation for \textit{instance-specific} mixing of auxiliary objectives in supervised learning. This is, to our knowledge, the first proposal of its kind  ({\em c.f.} Section~\ref{sec:adaptiveLossMixing}). \\ 
     \noindent 2) \textbf{\model\  in KD settings:} We explore a range of settings in Knowledge Distillation (KD), including vanilla KD, multi-teacher KD, and early-stopping, showing significant gains over and above SOTA KD approaches  in these settings ({\em c.f.} Section~\ref{sec:distillation}). \\
    \noindent 3) \textbf{\model\  in rule-denoising setting with limited supervision:} We show how the problem of semi-supervised data programming can benefit from \model\, and report gains of 2-5\% on various datasets  ({\em c.f.} Section~\ref{sec:RuleMixedLimited}). \\
   

\section{Related work}


\noindent \textbf{Knowledge distillation (KD)}
KD~\cite{hinton2015distilling} in a supervised learning setting trains a `student' model to mimic the outputs of a larger, pre-trained `teacher' model instead of directly training on the supervised signal. The efficacy of KD can be limited by teacher accuracy (see~\cite{pmlr-v139-menon21a} for some theoretical results), and student representational capacity, among other factors. Interestingly, early stopped teacher models aid better in training the student models~\cite{cho2019efficacy}; however, identifying the best possible teacher requires repeating the distillation process multiple times on the student model. To bridge the representational gap between the teacher and the students, Teacher Assistants (TA) or intermediate models were introduced~\cite{mirzadeh2019improved}, and were improved by a stochastic approach (DGKD~\cite{son2021densely}) for simultaneously training all intermediate models with occasional model dropout. In~\cite{liu2020adaptive}, multiple teacher networks are used with an intermediate knowledge transfer step using latent distillation. {\em All these works attempt to improve KD efficacy in cases in which there is a large gap between the teacher and student model as in the case  presented by us in Figure~\ref{fig:motivatio_kd}. However these methods require us to independently train additional models, in contrast to our work wherein we strategically mix loss components.} 

\noindent \textbf{Instance-Specific Learning:} A significant amount of past literature has explored instance-specific learning, for instance instance-specific temperature parameters in supervised learning~\cite{saxena2019data}. Other closely related work~\cite{algan2021meta, vyas2020learning} learns a per-instance \textit{label uncertainty} parameter to account for potential label noise. In the distillation setting, too,~\citet{zhao2021knowledge} demonstrate the benefits of  learning an instance-level sequence (or curriculum) on training samples. \citet{castells2020superloss} propose a task-agnostic per-sample loss-function representing the reliability of each prediction. Other recent works such as~\cite{l2r,meta_net,comm}, use validation set based meta learning to learn instance-specific weights to improve robustness.  
{\em The novelty of our work is that we seek task-agnostic, per-sample, loss mixing coefficients,  specifically for effective learning over multiple losses. }

\noindent \textbf{Bi-level Optimization and Meta-Learning}: Prior work~\cite{jenni2018deep,bengio2000gradient,domke2012generic}  has explored learning network hyper-parameters via solving a two-level optimization problem--one on the base-task and another on an external model-selection or meta-task, often on validation data. These algorithms are  similar in spirit to the \textit{learning to learn} literature, typically in  multi-task contexts~\cite{finn2017model,nichol2018first,hospedales2020meta,vyas2020learning}. Typical approaches aim to learn a ``meta-''algorithm which can generalize across tasks by mimicking the  test dynamics (sampling test tasks, in addition to test data, for measuring and optimizing loss) during training \cite{hospedales2020meta}. Although this literature, too, employs nested optimization objectives, {\em it differs from our work in that we wish to improve generalization within a single task, rather than across tasks.}

\noindent \textbf{Training with auxiliary tasks}: Information from auxiliary tasks are used to improve the main task in methods like ~\cite{lin2019adaptive,navon2021auxiliary} learn to reweigh auxiliary tasks to improve performance on the main task. \citet{guo2018dynamic}  construct a dynamic curriculum by weighing individual auxiliary tasks.  Similarly, \citet{NEURIPS2020_4f87658e} weigh auxiliary tasks to perform learning in a limited labeled data setting. The aforementioned approaches focus on unifying several losses into a single coherent loss {\em whereas our focus is on instance-wise contribution of the loss components. }

\section{\model: Adaptive Mixing of Auxiliary Losses }\label{sec:adaptiveLossMixing}

We consider the scenarios in which there are two or more loss terms participating in a supervised learning setting.  The loss functions we consider adhere to the form specified in Eq.~\ref{eqdis1}, where there is a primary objective and $K$ auxiliary objectives. 
\vspace{-1em}
\begin{align}
\Lcal = \lambda_p*\Lcal_p +\sum_{k=1}^K \lambda_{a_{k}} \mathcal{L}_{a_{k}}
\label{eqdis1}
\end{align}
Here, $\Lcal_p$ and $\Lcal_{a_i}$ respectively are the primary and auxiliary loss objectives. While this formulation is general, in this paper, we explicate the formulation in two different settings -- knowledge distillation (Section~\ref{sec:distillation}), and rule-denoising (Section~\ref{sec:RuleMixedLimited}). In these settings, we begin with a labeled dataset $\Dcal =\{(\xb_i,y_i)\}^N_{i=1}$ with instances $\xb_i$ and categorical labels $y_i$ and an unlabelled dataset  $\Ucal =\{(\xb_i)\}^{N+M}_{i=N}$   only with instances $\xb_i$. Note that, in the knowledge distillation setting, $\Ucal$ will be empty and in the case of rule-based denoising setting
$N << M$.  
Our main proposal is to modify the objective in Eq.~\eqref{eqdis1} so that loss-mixing coefficients ($\Lambda$) are instance-specific. Formally, we modify the loss function in Eq.~\eqref{eqdis1} as follows: 
\begin{align}
      \mathcal{L}(\theta,\mathbf{\Lambda}) =  \sum_i\left( \lambda_{p_i}\mathcal{L}_p\left(y_i, \xb_i|\theta\right)  + \sum_{k=1}^K \lambda_{a_{k,i}} \mathcal{L}_{a_{k,i}}\right)
    \label{eq:instance}
\end{align}

\begin{algorithm}[t]
  \caption{Algorithm for learning $\lambda$s via meta learning}
  \label{alg:algMeta}
  \begin{algorithmic}[1]
    \REQUIRE Training data ${\mathcal U}$, Validation data ${\mathcal V}$, $\theta^{(0)}$ model parameters initialization, $\tau$ Temperature,$\eta$: learning rate, $\eta_{\lambda}$: learning rate for updating $\lambda$.
    \REQUIRE $LL_{CE}$ Primary Supervised Loss, $L_{a}$  auxiliary loss, max iterations $T$
    \STATE Initialize model parameters $\theta^{(0)}$ and $\lambda^0_p,\lambda^0_{a_1},\lambda^0_{a_2},\cdots,\lambda^0_{a_K}$.
    \FOR {$t \in \{0,\dots,T\}$}
        \STATE Update $\theta^{t+1}$ by Eq. \eqref{eq:final}.
        \IF{t \% L == 0}
            \STATE {\small ${\xb^{train}, y^{train}} \xleftarrow{} \texttt{SampleMiniBatch}({\mathcal U})$}
            \STATE ${\xb^{val}, y^{val}} \xleftarrow{} \texttt{SampleMiniBatch}({\mathcal V})$
            \STATE Compute one step update for model parameters as function of $\lambda_{p}^{\lfloor\frac{t}{L}\rfloor},\lambda_{a_{1}}^{\lfloor\frac{t}{L}\rfloor},\lambda_{a_{2}}^{\lfloor\frac{t}{L}\rfloor}, \cdots, \lambda_{a_{K}}^{\lfloor\frac{t}{L}\rfloor}$ by Eq.\eqref{eqdis2}.
    
           \STATE Update $\Lambda^{\lfloor\frac{t}{L}\rfloor}$ by Eq.\eqref{eq:meta_up}.
        \ENDIF
    \ENDFOR    

  \end{algorithmic}
\end{algorithm} 

 
\begin{figure*}[t]
\centering
\includegraphics[width = 0.8\linewidth, height=.7cm,trim={0cm .5cm 0 0},clip]{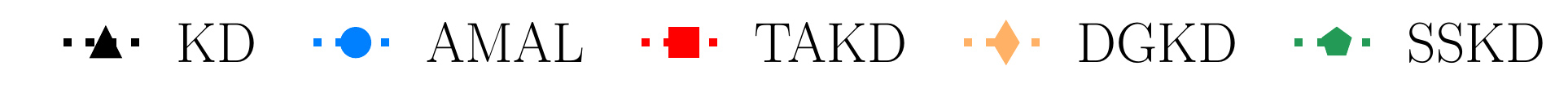}
\centering
\hspace{-0.6cm}
\begin{subfigure}[b]{0.32\textwidth}
\centering
\includegraphics[width=\textwidth, height=4cm]{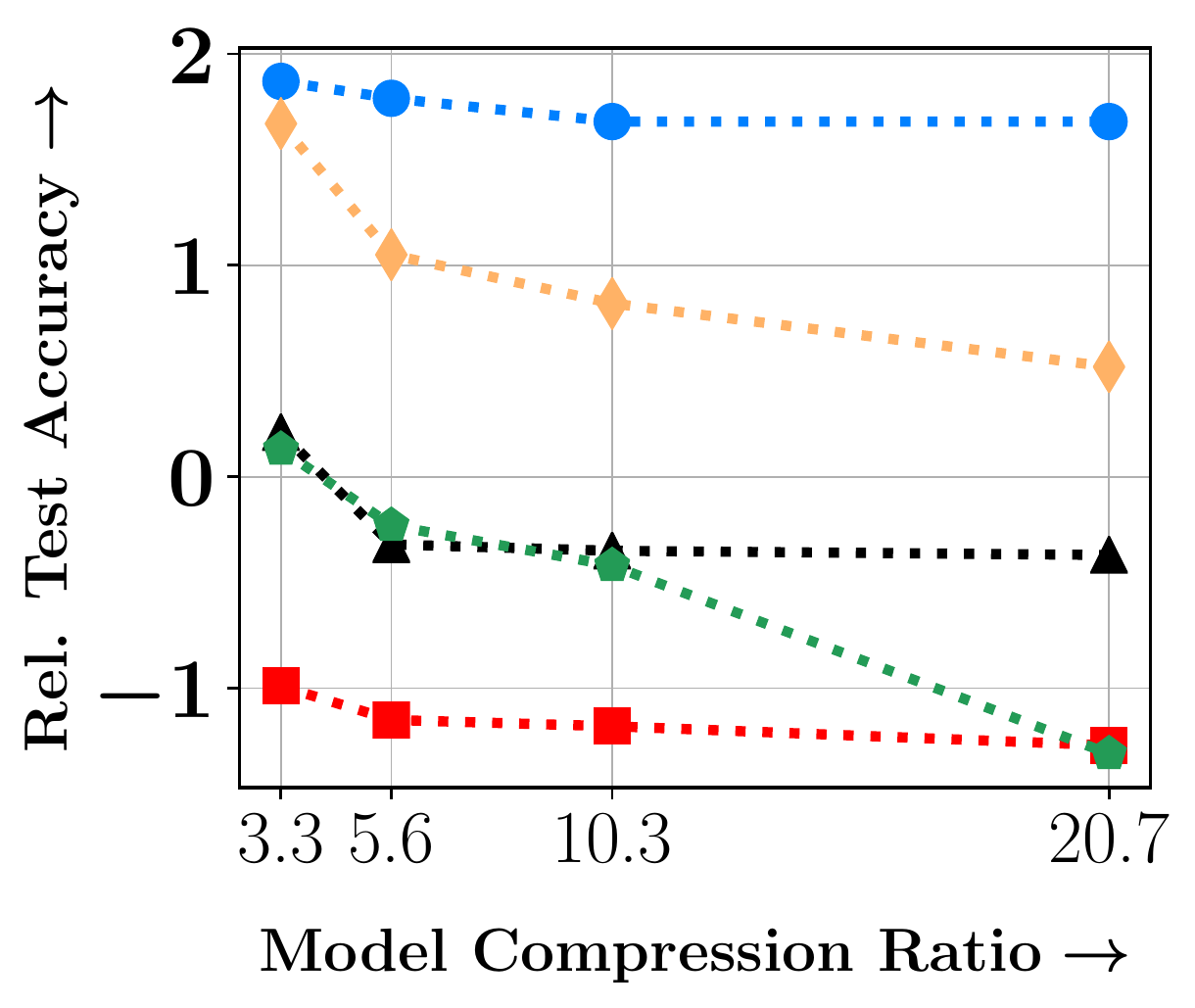}
\caption*{$\underbracket[1pt][1.0mm]{\hspace{4.5cm}}_{\substack{\vspace{-4.0mm}\\
\colorbox{white}{(a) \scriptsize CIFAR100}}}$}
\phantomcaption
\label{fig:CIFAR100}
\end{subfigure}
\begin{subfigure}[b]{0.32\textwidth}
\centering
\includegraphics[width=\textwidth, height=4cm]{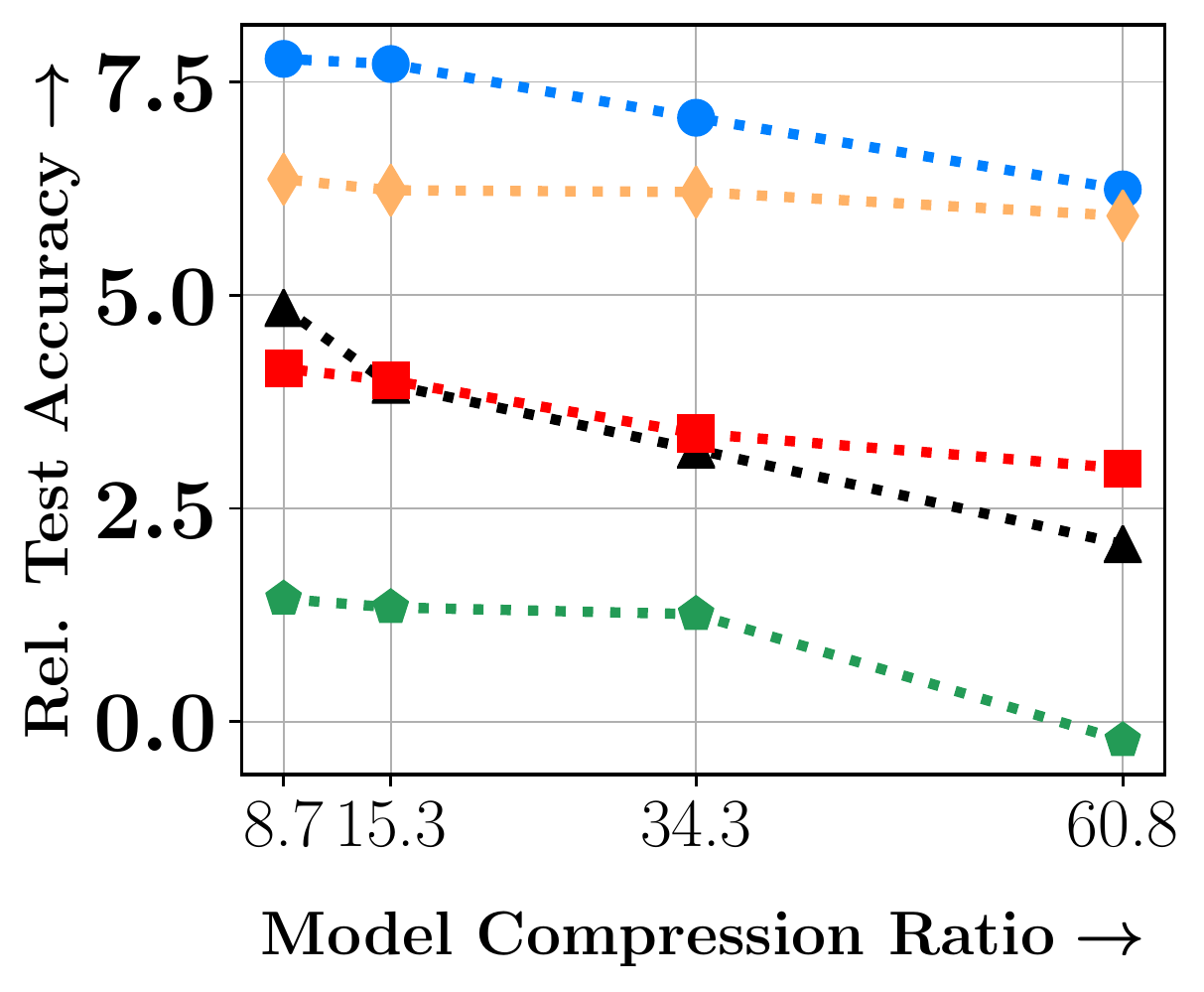}
 \caption*{$\underbracket[1pt][1.0mm]{\hspace{4.5cm}}_{\substack{\vspace{-4.0mm}\\ \colorbox{white}{(b) \scriptsize Stanford Cars}}}$}
\label{fig:Cars}
\end{subfigure}
\begin{subfigure}[b]{0.32\textwidth}
\centering
\includegraphics[width=\textwidth, height=4cm]{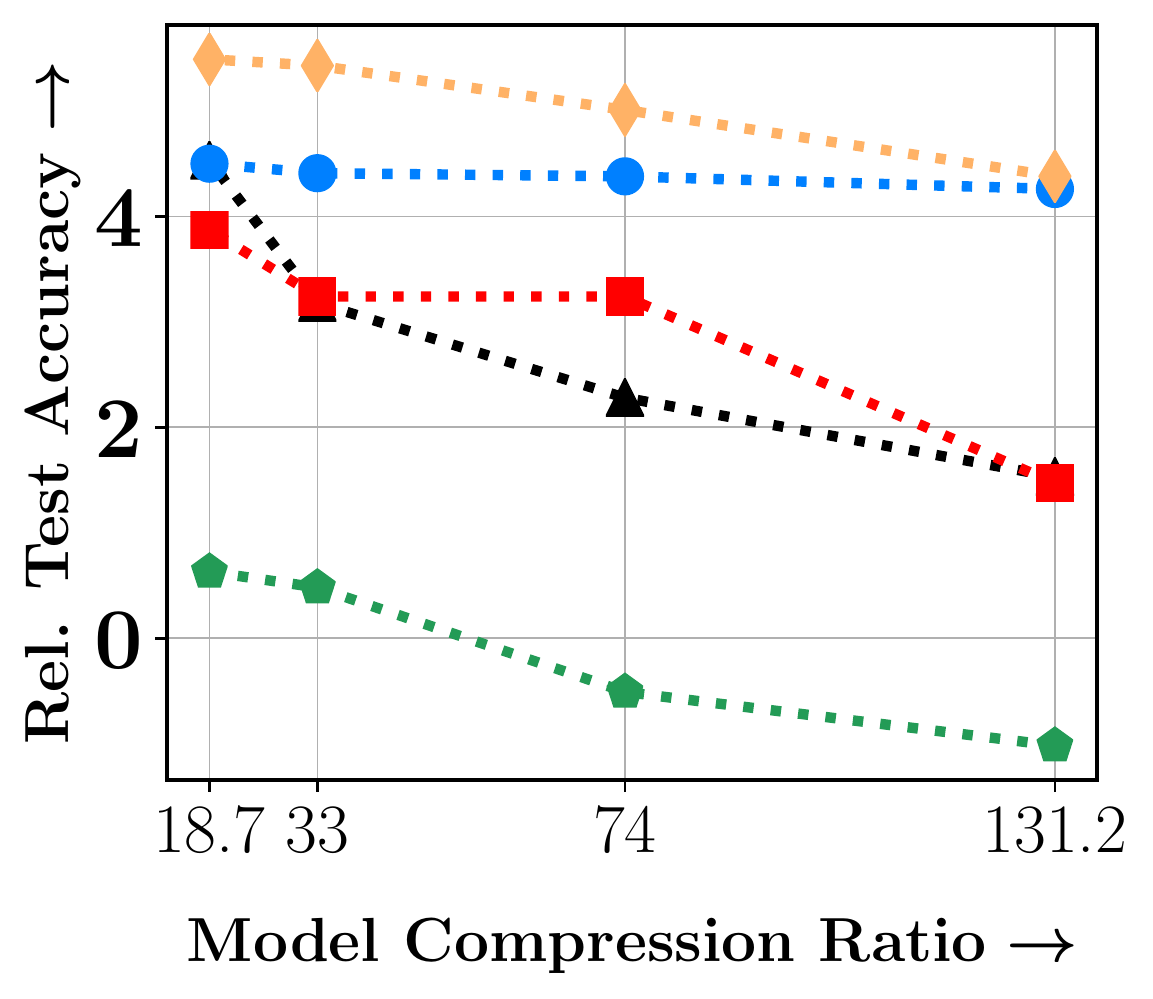}
\caption*{$\underbracket[1pt][1.0mm]{\hspace{4.5cm}}_{\substack{\vspace{-4.0mm}\\
 \colorbox{white}{(c) \scriptsize FGVC-Aircraft}}}$}
\phantomcaption
\label{fig:plane}
\end{subfigure}

\caption{ KD performed  with  ResNet (8,20,32,56,110) models on CIFAR100 in subfigure (a), Wide Residual Networks (WRN-16-1, WRN-16-3,WRN-16-4,WRN-16-6,WRN-16-8) on Stanford Cars in subfigure (b) and with Wide Residual Networks (WRN-16-3,WRN-16-4,WRN-16-6,WRN-16-8) as teachers and Resnet8 as student on FGVC-Aircraft in subfigure (c).  \model\ consistently outperforms various state-of-the-art methods when the model compression ratio is higher.}
\label{fig:kd_experiments}
\end{figure*}

Note that formulation in Eq.~\eqref{eq:instance} is a generalization of Eq.~\eqref{eqdis1}, with an instance-specific value of mixing parameters $\Lambda = \{\lambda_{p},\lambda_{a_{1}},\lambda_{a_{2}}, \cdots, \lambda_{a_{K}}\}$ corresponding to the $i^{th}$ training instance $\xb_i$. Jointly optimizing the objective in Eq.~\eqref{eq:instance} with respect to both sets of parameters $\theta, {\bf \Lambda}$ on the training dataset alone can lead to severe overfitting. To mitigate this risk, we instead  attempt to solve the bi-level minimization problem in Eq.~\eqref{nested-equation} using a meta-learning procedure:

\vspace{-1.2em}
\begin{align}
    \overbrace{\underset{{\Lambda}}{\operatorname{argmin\hspace{0.7mm}}} \mathcal{L}_{CE}\big(\underbrace{\underset{\theta}{\operatorname{argmin\hspace{0.7mm}}} \mathcal{L}( \theta,\Lambda)}_{inner-level}, {\mathcal V}\big)}^{outer-level}
    \label{nested-equation}
\end{align}
 By solving the inner level minimization, we wish to obtain  model parameters $\theta$ that minimise the objective in Eq.~\eqref{eq:instance}. The outer minimization yields $\lambda$s such that the standard cross-entropy loss is minimised on the validation set $\Vcal$. This problem is a bi-level optimisation problem since model parameters $\theta$ are dependent on $\Lambda$ and computation of $\Lambda$ is dependent on model parameters $\theta$ as shown in Eq.\eqref{nested-equation}.
 
Since the inner optimisation problem cannot be solved in a closed form in Eq.~\eqref{nested-equation}, we need to make some approximations in order to solve the optimization problem efficiently.  We take an iterative approach,  simultaneously updating the   optimal model parameters $\theta$ and appropriate $\Lambda$ in alternating steps as described in the Algorithm\ref{alg:algMeta}.  We first update the model parameters by sampling a mini-batch with $n$ instances from the training set, and simulating a one step look-ahead SGD update  for the loss in Eq.\eqref{eq:instance} on model parameters $(\theta^{t})$ as a function of $\Lambda^t$, resulting in Eq.~\eqref{eqdis2}, with $L$ being a hyperparameter governing how often the lambda values are updated.
\vspace{-1em}
\begin{align}
        \label{eqdis2} 
      \hat{\theta^{t}}\left(\Lambda^{\lfloor\frac{t}{L}\rfloor} \right) = \theta^{t} - \frac{\eta}{n}\sum_{i=1}^{n} \nabla_{\theta^{t}} \mathcal L_{i}(\theta^{t}(\Lambda_i^{\lfloor\frac{t}{L}\rfloor}) )
\end{align}
Using the approximate model parameters obtained using the one step look-ahead SGD update, the outer optimization problem is solved as,
\vspace{-.1em}
\begin{align}
  \label{eq:meta} 
  \nabla_{\Lambda_i^{\lfloor\frac{t}{L}\rfloor} }&\mathcal L_{CE}(\hat{\theta^{t}},{\mathcal V})  \\&= - \frac{\eta}{n}.\nabla_{\hat{\theta^{t}}}\mathcal L_{CE}(\hat{\theta^{t}},{\mathcal V}). \nabla_{\Lambda_i^{\lfloor\frac{t}{L}\rfloor} } \nabla_{\theta^{t}}\mathcal L_{i}^T
   \nonumber
\end{align}
We derive gradients for each of loss objective in Appendix \ref{add_imp}. Using the meta-gradient in Eq.\eqref{eq:meta} we update the $\lambda$s  for each of the training samples using the first order gradient update rule as, 
\vspace{-.1em}
\begin{align}
      \Lambda_i^{\lfloor\frac{t}{L}\rfloor +1 } = \Lambda_i^{\lfloor\frac{t}{L}\rfloor} - \eta_{\lambda} \nabla_{\Lambda_i^{\lfloor\frac{t}{L}\rfloor}} \mathcal L_{CE}(\hat{\theta^{t}},{\mathcal V}) 
    \label{eq:meta_up}
\end{align}

Here, $\eta_{\Lambda}$ is the learning rate for mixing parameters. We update $\Lambda$ values every L epochs. The updated  $\Lambda_i^{\lfloor\frac{t}{L}\rfloor+1}$ values are then used to update the model parameters as,  
\vspace{-.1em}
\begin{align}
    \label{eq:final}
     \theta^{t+1} = \theta^{t} - \frac{\eta}{n}\sum_{i=1}^{n} \nabla_{\theta^{t}}  \mathcal L_{i}(\theta^{t}(\Lambda_{i}^{\lfloor\frac{t}{L}\rfloor+1}))
\end{align}


In Appendix \ref{convergence}, we show theoretically that our method converges to the optima of both the validation and training loss functions under some mild conditions.

\subsection{Speeding up \model{}}

We borrow two important implementation schemes from few of the recent subset selection techniques~\cite{killamsetty2021glister,pmlr-v139-killamsetty21a} to streamline mixing parameter updates in \model. Firstly, instead of using the complete high dimensional loss gradient associated with modern deep neural networks we only consider last-layer gradient of a network. This helps in reducing both computation time and memory in both  the one step update (Eq.~\eqref{eqdis2}) and computation of the meta-gradient (Eq.~\eqref{eq:meta}). Similarly, the proposal to update $\Lambda$ only after $L$ epochs also reduces the computation time significantly. Bi-level optimisation solved with these tricks has been shown to yield significant speedup~\cite{killamsetty2021glister} and with minimal loss in performance. Thus, training with \model{} introduces negligible overhead. 

\section{Two Application Scenarios for \model}\label{sec:application}
In this Section, we present two application scenarios for \model, described in the previous Section~\ref{sec:adaptiveLossMixing}, {\em viz.}, knowledge distillation \ref{sec:distillation} and learning with limited supervision and rule-denoising in Subsection~\ref{sec:RuleMixedLimited}. 

\subsection{Knowledge distillation}\label{sec:distillation}

Any (student) model having output logits  as $a^{(S)} = \texttt{StudentModel}(\xb)$, is traditionally trained by optimizing a cross-entropy based loss $\mathcal{L}_s$ defined as follows: 
\begin{align}
      \mathcal{L}_s = L_{CE}\big(\texttt{softmax}(a^{(S)}),y\big)
    \label{eqdis0}
\end{align}
Let us say we have access to a pretrained teacher model (typically of higher learning capacity) which outputs the logits $a^{(T)} = \texttt{TeacherModel}(\xb)$. We can frame a \textit{teacher matching} objective for the student as a KL-divergence between the predictions of the student and the teacher: 
\begin{align}
      \mathcal{L}_{KD} = 	\tau^2 KL\big(y^{(S)},y^{(T)}\big) \label{eq2}
\end{align}
Then the training of the student model can be performed using both the \textit{teacher matching} objective and the traditional cross entropy loss  as,
\vspace{-.1em}
{\small
\begin{align}
      \mathcal{L}_{student}(\theta,\mathbf{\lambda}) =  \sum_i (1 - \lambda)\mathcal{L}_s\left(y_i, \xb_i|\theta\right) 
     + \lambda \mathcal{L}_{KD}\left(y_i^{(S)}, y_i^{(T)}\right)
    \label{eq:distil_std}
\end{align}}

This is the standard knowledge distillation loss, in which a temperature parameter $\tau$ is typically used to control the softening of the KD loss in Eq.~\eqref{eq2}; therefore we have $y^{(S)} = \texttt{softmax}\big(\frac{a^{(S)}}{\tau}\big)$ and $y^{(T)} = \texttt{softmax}\big(\frac{a^{(T)}}{\tau}\big)$. We change this objective to match \model's objective as, 
{\small
\begin{align}
      \mathcal{L}_{student}(\theta,\mathbf{\Lambda}) =  \sum_i \lambda_{p_i}\mathcal{L}_s\left(y_i, \xb_i|\theta\right) 
     + \lambda_{a{1_i}} \mathcal{L}_{KD}\left(y_i^{(S)}, y_i^{(T)}\right)
    \label{eq:distil}
\end{align}}

{Clearly, here $\Lcal_p$ would be $\Lcal_s$ and $\Lcal_a$ would be $\Lcal_{KD}$. We present the results of applying \model\ to adaptively mix these losses in Section~\ref{sec:kdresults}}. \model{} can be extended to settings where distillation is performed with multiple teachers such as DGKD~\cite{son2021densely}. We present details with additional experiments in Section  \ref{sec:multi}.

\subsection{Learning with limited supervision and rule-denoising \label{sec:RuleMixedLimited}}
Several rule-denoising approaches ~\cite{maheshwari-etal-2021-semi, awasthi2020learning, cage, ratner2017snorkel} encode multiple heuristics in the form of rules (or labeling functions) to weakly associate labels with instances. These weak labels are aggregated to determine the probability of the correct labels using generative models~\cite{cage, ratner2017snorkel} without requiring labeled data. In contrast, recent approaches~\cite{maheshwari-etal-2021-semi,karamanolakis2021self, awasthi2020learning,ren2020denoising, l2r} assume that a small labeled dataset is available in conjunction with the noisy rules.
Motivated by the success of rule denoising approaches, we propose adaptive loss mixing to leverage a small labeled set while being trained in a joint manner. We directly adopt the model and loss formulations from the most recent of these approaches~\cite{maheshwari-etal-2021-semi}, since it performs consistently better than the previous ones~\cite{maclaurin2015gradient, awasthi2020learning,ren2020denoising,l2r} (see Section~\ref{sec:LimSuper}).  


Our setting borrowed from SPEAR~\cite{maheshwari-etal-2021-semi} is as follows: In addition to the setting described in Section \ref{sec:adaptiveLossMixing}, we also have access to $m$ rules or labelling functions (LF) $lf_1$ to $lf_m$. We modify $\Dcal$ to be $\Dcal' =\{(\xb_i,y_i,l_i)\}^N_{i=1}$ and $\Ucal$ to be $\Ucal' =\{(\xb_i,l_i)\}^{N+M}_{i=N}$, where $l_i = (l_{i1},l_{i2},\cdot,l_{im})$ is a boolean vector with $l_{ij}=1$ if the corresponding $j^{th}$ LF  is activated on example $\xb_i$ and $l_{ij}=0$ otherwise. Exactly as per~\cite{maheshwari-etal-2021-semi}, our model is a blend of the feature-based classification model  $f_\theta(\bfx)$ and the rule-based model $P_{\phi}(\bfl_i, y)$. We have two types of supervision in our joint objective. First, we have access to  $y$ for  the labeled instances $\Dcal'$ and to $l_{ij}$ for all the labeled as well as unlabeled instances $\Dcal' \bigcup \Ucal'$. We measure the loss of $P_\theta$ and $P_\phi$ on the small labeled set $\Dcal'$ through standard cross-entropy. Second, we model interaction between  $P_\theta$ and $P_\phi$ on the union of labeled and unlabeled sets. Intuitively, the rule denoising model $P_\phi$ learns with respect to the clean labeled set $\Dcal'$ and simultaneously  provides labels over $\Ucal$ that can be used to train the feature model $P_\theta(y|x)$. We want both the models to agree in their predictions over the union $\Dcal' \bigcup \Ucal'$ (Please refer to Supplementary Section E for details about individual loss components.)

\section{Results}

In this section, we present results for the two application scenarios for \model{} as outlined in Section~\ref{sec:application}.

\subsection{Results with Knowledge Distillation} \label{sec:kdresults}
In this section, we report a range of experimental results from the knowledge distillation (KD) scenario as described in Section~\ref{sec:distillation}. 
We performed a range of experiments comparing \model\ against several SOTA knowledge distillation approaches on several real-world datasets, with a special focus on those settings wherein we found the gap between the teacher and student models to be large. 

\noindent \textbf{Datasets}
The datasets in our experiments include CIFAR100 ~\cite{Krizhevsky09learningmultiple},  Stanford Cars~\cite{KrauseStarkDengFei-Fei_3DRR2013}and FGVC-Aircraft~\cite{maji13fine-grained}; characteristics of the datasets are summarized in Table~\ref{tab:textdatasplits} in the Appendix. For the CIFAR datasets we used the standard RGB images of size $32\times 32$, whereas for the other datasets we used RGB images of size $96 \times 96$.

\noindent \textbf{Model  architecture  and  experimental  setup} We explored two families of models, {\em viz.}, (i) Wide Residual Networks (WRN-16-1, WRN-16-3,WRN-16-4,WRN-16-6,WRN-16-8)~\cite{Zagoruyko2016WRN}, and (ii)  ResNet (8,20,32,56,110) models ~\cite{he2016deep} to show the effectiveness of our method across the different model families. We also perform a distillation  on Resnet8 with WRN-16-3,WRN-16-4,WRN-16-6 and WRN-16-8 as teachers to show the effect of our technique in the cross-model distillation. 

For datasets without pre-specified validation sets, we split the original training set into new train (90\%) and validation sets (10\%) (see Table~\ref{tab:textdatasplits} for details). Training consisted of SGD optimization with an initial learning rate of 0.05,  momentum of 0.9, and weight decay of 5e-4. We divided the learning rate by 0.1 on epochs 150, 180 and 210 and trained for a total of 240 epochs.



\noindent \textbf{Effect of optimal $\lambda$s on Knowledge Distillation} In the first experiment, we examine effective transfer of learned knowledge from various teachers to a student model which has fewer parameters. We compares test accuracies obtained with KD, \model\ , TAKD~\cite{mirzadeh2019improved} and DGKD~\cite{son2021densely} and SSKD \cite{xu2020knowledge}. TAKD takes taking multiple KD training hops, with each step reducing the model complexity from teacher to student by a small amount. DGKD introduces all the intermediate teachers in a single KD training step using a single $\lambda$, across all training instances, for each teacher.  In addition, stochastic DGKD was proposed where a subset of teachers is introduced at each training step, determined by a binomial (hyperparamter) variable.

\noindent \textbf{Additional experimental setup} We perform KD with  ResNet (20,32,56,110) as teacher and  ResNet8 as student models on CIFAR100, Wide Residual Networks (WRN-16-3,WRN-16-4,WRN-16-6,WRN-16-8) as teacher and WRN-16-1 as student models on Stanford Cars and with Wide Residual Networks (WRN-16-3,WRN-16-4,WRN-16-6,WRN-16-8) as teachers and Resnet8 as student on FGVC-Aircraft. For TAKD and DGKD we use ResNet14 for CIFAR100 and WRN-16-2 for Stanford Cars and FGVC-Aircraft as teaching assistant models. In all our knowledge distillation experiments we use temperature $\tau = 4$ and $\lambda_a =0.9$ (weights associated with KD loss) except in case of \model. For DGKD we use set the binomial variable to be $0.75$, best reported in the paper. 


\noindent Figure~\ref{fig:kd_experiments} shows that \model{} consistently outperforms other techniques when a much smaller model learns from large teacher model (CIFAR100, Stanford Cars) and is comparable to DGKD in FGVC-Aircraft dataset. The figure shows  plot relative test accuracies (w.r.t. non-KD students) vs model compression ratio\footnote{We define model compression ratio as (no. of learnable parameters in teacher model)/(no. of learnable parameters in student model);  higher is better}.Interestingly, methods such as KD, SSKD and TAKD actually perform worse than training a student model with standard cross entropy loss. This observation is consistent with ~\cite{cho2019efficacy}, where authors claim KD may fail if the student is too weak. This problem gets worse when techniques such as SSKD bring even more additional information for the student model to learn. TAKD tries to address this issue by bring in teaching assistant model, which have already gone through with knowledge distillation from the teacher model. However, this also transfer errors from the higher level to the lower level models \cite{son2021densely}. It is important to note that \model\ doesn't require any additional intermediate model to be trained like TAKD and DGKD and therefore has a lesser memory footprint and training time. 

\subsubsection{Knowledge Distillation in presence of noise}\label{sec:KD_noise}

\begin{figure}[t]
\centering
\includegraphics[width = \linewidth,height=.6cm ,trim={0cm .5cm 0 0},clip]{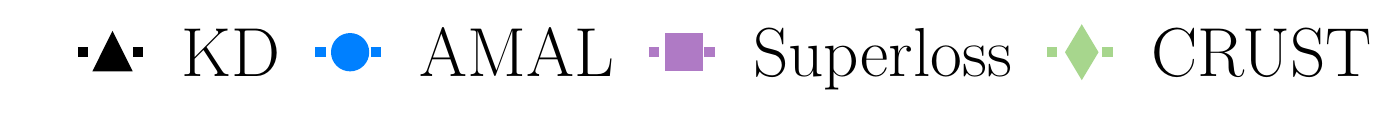}
\centering
\hspace{-0.6cm}
\begin{subfigure}[b]{0.49\linewidth}
\centering
\includegraphics[width=\linewidth]{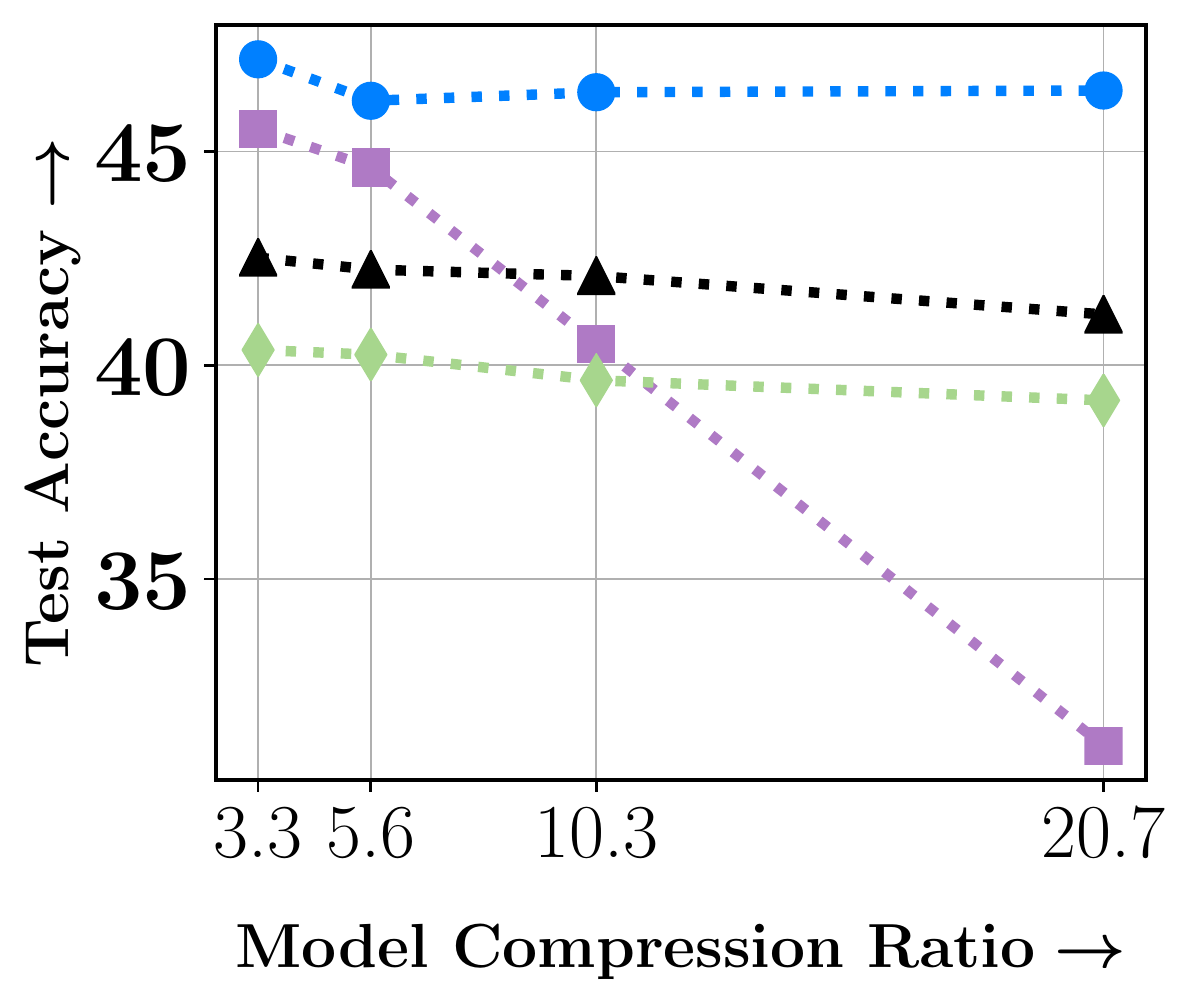}
\caption*{$\underbracket[1pt][1.0mm]{\hspace{3.8cm}}_{\substack{\vspace{-4.0mm}\\
\colorbox{white}{(a) \scriptsize 40\% noise}}}$}
\phantomcaption
\end{subfigure}
\begin{subfigure}[b]{0.49\linewidth}
\centering
\includegraphics[width=\linewidth]{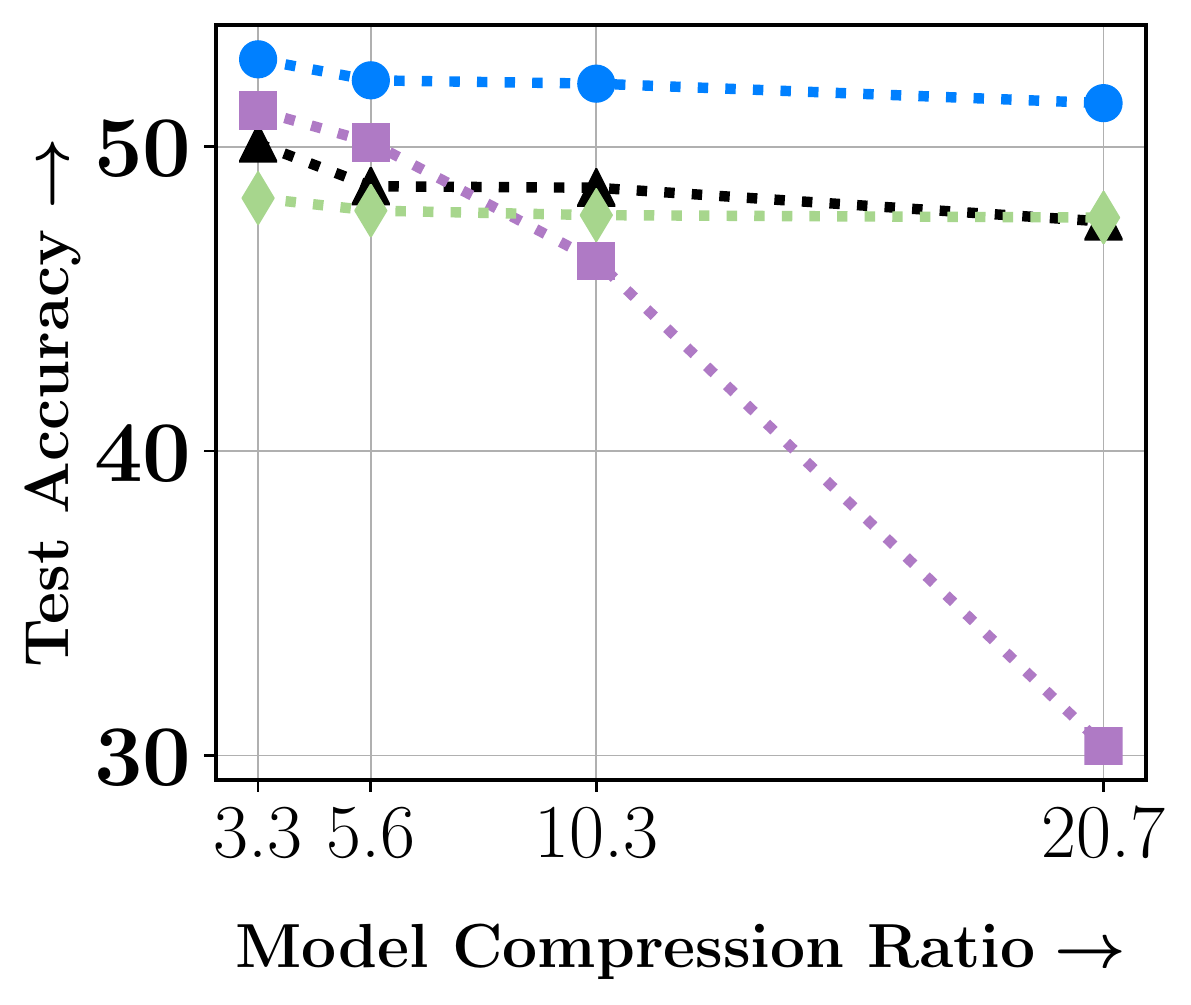}
 \caption*{$\underbracket[1pt][1.0mm]{\hspace{3.8cm}}_{\substack{\vspace{-4.0mm}\\ \colorbox{white}{(b) \scriptsize 60\% noise}}}$}
\end{subfigure}
\caption{Test performance obtained after performing knowledge distillation with ResNet8 as the student and Resnet (20,32,56,110) as the teachers with CIFAR100 dataset corrupted with 40\% and 60\% label noise.}
\label{fig:noise_distill}
\end{figure}

As \model{} performs instance wise mixing of loss components, noise filtering in knowledge distillation  (with two loss components) is an appropriate use case. We perform knowledge distillation with CIFAR100 dataset with n\% labels randomly changed to a wrong label. We continue using the ResNet (8,20,32,56,110) model with ResNet8 being the student model. We present test accuracies obtained while training with 40\% and 60\% label noise in Figure \ref{fig:noise_distill}. We compare against two loss agnostic robust learning techniques {\em viz.} (i) \textbf{Superloss} \cite{castells2020superloss}: It is curriculum learning based approach which dynamically assigns weights to each instance to perform robust learning.(ii) \textbf{CRUST}  \cite{crust}: It selects a noise free subset of data points which approximates the low-rank Jacobian matrix.

Figure \ref{fig:noise_distill} we see that \model{} achieves best performance which could be explained by the mixing parameters'($\Lambda$) distribution presented in Figure \ref{distill_diff}. \model{} identifies importance of learning form cross entropy based loss for the clean points and learning from KD loss for noisy points. However, CRUST as it selects a subset selection it can't take advantage of both the losses. Superloss, on the other enjoys performance improvement over KD  for smaller model compression ratios. However, superloss's performance drops significantly for higher compression ratios as it doesn't perform any kind of mixing. We present more analysis on $\Lambda$ values learnt in Appendix \ref{lam_analysis}.

\subsection{Connection to a Coreset}

\begin{table}[t]
    \centering
    \begin{tabular}{c| c } \hline \hline
    Method  & Test Accuracy    \\ [0.7ex]\hline
 {Complete data (skyline)} & 66.43 \\[0.7ex] \hline
 {Random}& 44.92 \\ [0.7ex]\hline 
 {Sampled according to $\lambda_p^2+\lambda_a^2$}& 45.5 \\ [0.7ex]\hline 
  {Sampled according to $|\lambda_p-\lambda_a|$}& 46.28   \\ [0.7ex]\hline
  {Sampled according to $\frac{\lambda_a}{\lambda_p}$} & 46.31 \\[0.7ex] \hline
 \hline 
 \hspace{1em}
    \end{tabular}
    \caption{Test accuracies obtained after training  with 20\% subset obtained using various strategies using the WRN-16-1 model on the CIFAR100 dataset. We perform training  only with the CE loss.}
    \label{tab:coreset}
    
\end{table}

Since, \model{} controls the contribution of each of the instances in training a model by weighting each of the points loss functions. We try to understand the significance of the weights associated with each data point with a  coreset based experiment. Coreset selection has become popular in recent times where a subset of training points are used to train a model from scratch. Based on the final $\lambda_p$(weighted associated with the CE loss) and $\lambda_a$(weighted associated with the KD loss) values while training WRN-16-1 model using WRN-16-8 as a teacher model on CIFAR100 dataset, we derive a probability of selection $p_i$ for each point $i$ in the training set as, 

\begin{enumerate}
    \item $p_i \propto \lambda_{p_i}^2+\lambda_{a_i}^2$, here we pick points with maximum weights as they would contribute maximum to the model training 
    \item $p_i \propto |\lambda_{p_i}-\lambda_{a_i}|$, here we pick points that should be preferably learnt with only one of the losses
    \item $p_i \propto \frac{\lambda_{a_i}}{\lambda_{p_i}}$, here we pick points that should be preferably learnt with only KD loss
\end{enumerate}

In Table~\ref{tab:coreset} we present the  test accuracies obtained on training WRN-16-1 with the coresets obtained when sampled using the corresponding probabilities. We also present the result of training the same model with randomly (sampled with uniform distribution) obtained subset. We train with subsets of 20\% size of the original training data and train with only the CE loss. Clearly, the points that have higher weights have maximum information. More, specifically the points that require a teacher model's assistance and cannot be learned using the ground truth seem to have the most information and therefore coreset formed using 3 performs the best.


In Supplementary Section \ref{app:experiments},  we report the use of the validation data in different forms to strengthen baseline, but all those efforts either weakened or did not add any value to the existing baselines.  

\subsection{\model\ with limited supervision and rule-denoising} \label{sec:LimSuper}

\begin{table}[H]
\centering
\begin{tabular}{@{}lccc@{}}
\toprule
           & SMS   & IMDB  & YouTube \\ \midrule
Only-L     & 91.45 {\scriptsize (1.3)} & 77.35 {\scriptsize (1.5)} & 89.60 {\scriptsize (2.9)}   \\ \midrule
Imply Loss & +0.25 {\scriptsize (1)}  & -1.47 {\scriptsize (1.8)} & +2.70  {\scriptsize (0.8)}  \\
L2R        & -0.20 {\scriptsize (1.3)} & -2.18 {\scriptsize (1.4)}& +3.40 {\scriptsize (1.2)}    \\
MWN        & -0.10 {\scriptsize (1.2)} & -1.53 {\scriptsize (1.7)}& +3.70 {\scriptsize (1.5)}    \\
SPEAR  & -0.76 {\scriptsize (1.4)}  & -0.04 {\scriptsize (1)} & +4 {\scriptsize (1)}\\
\model{}        & \textbf{+1.53} {\scriptsize (0.9)}  & \textbf{+1.67} {\scriptsize (1.6)} & \textbf{+4.70} {\scriptsize (0.8)}    \\ \bottomrule
\end{tabular}%
\caption{Performance of our \model{} approach with  rule-based approaches Imply Loss, SPEAR, L2R and MWN. \model{} with fixed $\lambda=1$ corresponds to SPEAR. All numbers reported are gains over the baseline method (Only-L). All results are averaged over 5 random seed runs having different $\Lcal$ and $\Ucal$ set in each run. Numbers in brackets ‘()’ represent standard deviation of the original score.}
\label{tab:alm-spear}
\end{table}
In this section, we report our experimental results for the scenario of limited supervision combined with weak supervision from labeling functions (also referred to as semi-supervised data programming~\cite{maheshwari-etal-2021-semi}), as summarized in Section~\ref{sec:RuleMixedLimited}. 
\textbf{Datasets} We used three dataset in our experiments, namely, YouTube, SMS and IMDB. \textbf{YouTube} \cite{youtube} is a spam classification task over YouTube comments; \textbf{SMS } \cite{sms} is a binary spam classification containing 5574 documents; \textbf{IMDB} is a movie plot genre binary classification dataset.

In Table \ref{tab:alm-spear}, we compare our approach with the following approaches:
(1) \textbf{Only-$\Lcal$} : We train the classifier $P_\theta(y|x)$ only on the labeled data. To ensure fair comparison, we use the same classifier model for different datasets as mentioned in \cite{maheshwari-etal-2021-semi}. We choose this as a baseline and report gains over it. (2) \textbf{L2R}~\cite{l2r}: This is an online reweighting algorithm that leverage validation set to assign weights to examples based on gradient directions. It learns to re-weigh weak labels from domain specific rules and learn instance-specific weights via meta-learning.  (3) \textbf{Meta-Weight-Net(MWN)} ~\cite{meta_net} Trains a neural network assigns instantaneous weights. Neural network is trained to minimise validation set loss. However, weights are not learnt to mix losses in L2R and MWN. (4) \textbf{Imply Loss}~\cite{awasthi2020learning}: This is a rule-exemplar approach that jointly trains a rule denoising network and leverages \textit{exemplar}-based supervision for learning instance-specific and rule-specific weights. In addition, it also learns a classification model with a soft implication loss in a joint objective. (5) \textbf{SPEAR}~\cite{maheshwari-etal-2021-semi}: Finally, we compare with another rule-denoising approach that uses same objective as \model{} and is trained on both feature-based classifier and rule-classifier using a small labeled set. \model{} with all $\lambda$s fixed to $1$ (and not trainable) corresponds to SPEAR.

Our approach outperforms both rule-based and re-weighing approaches on all datasets. MWN, L2R and SPEAR perform worse than the baseline method (only-L) on SMS and IMDB dataset whereas Imply-Loss is marginally better on SMS. All approaches achieve better performance over the baseline method on YouTube dataset. However, \model{} consistently reports highest gains. Recall that SPEAR has the same objective as \model{} but without trainable $\lambda$s and all $\lambda$s fixed to 1. \model{} tries to identify instance-wise weighted combination of loss components so that the trained feature classification model performs better. Instance wise mixing is useful to identify the loss component from which a data point could be learned better and use of fixed weights prevents from understanding nuance of each data point.     

\section{Conclusion} 
In this paper we present two setting viz. rule-denoising setting with limited supervision and knowledge distillation (KD), where Adaptive Loss Mixing is useful. We present \model\ which via adaptive loss mixing extracts useful information from the limited supervision to prevent degradation of model learnt due to the presence of noisy rule. In knowledge distillation (KD) setting it titrates the teacher knowledge and ground truth label information through  an instance-specific combination of teacher-matching and ground supervision objectives to  learn student models that are more accurate. Our iterative approach is pivoted on solving a bi-level optimization problem in which the instance weights are learnt to minimize the CE loss on a held-out validation set whereas the model parameters are themselves estimated to minimize the weight-combined loss on the training dataset. Through extensive experiments on real-world datasets, we present how \model\ yields accuracy improvement and better generalization on a range of datasets in both the settings.


\section{Acknowledgements}
Durga Sivasubramanian is supported by the Prime Minister’s Research Fellowship. Ayush Maheshwari is supported by a Fellowship from Ekal Foundation (www.ekal.org). Ganesh Ramakrishnan is grateful to the IIT Bombay Institute Chair Professorship for their support and sponsorship. Prathosh acknowledges the support received via a faculty grant by Google Research India, for this work.  

\bibliography{refer}
\bibliographystyle{aaai23}

\onecolumn
\appendix

\section{Additional details on implementation} \label{add_imp}

\begin{figure}[H]
\centering
  \centering
  \includegraphics[width=.8\textwidth]{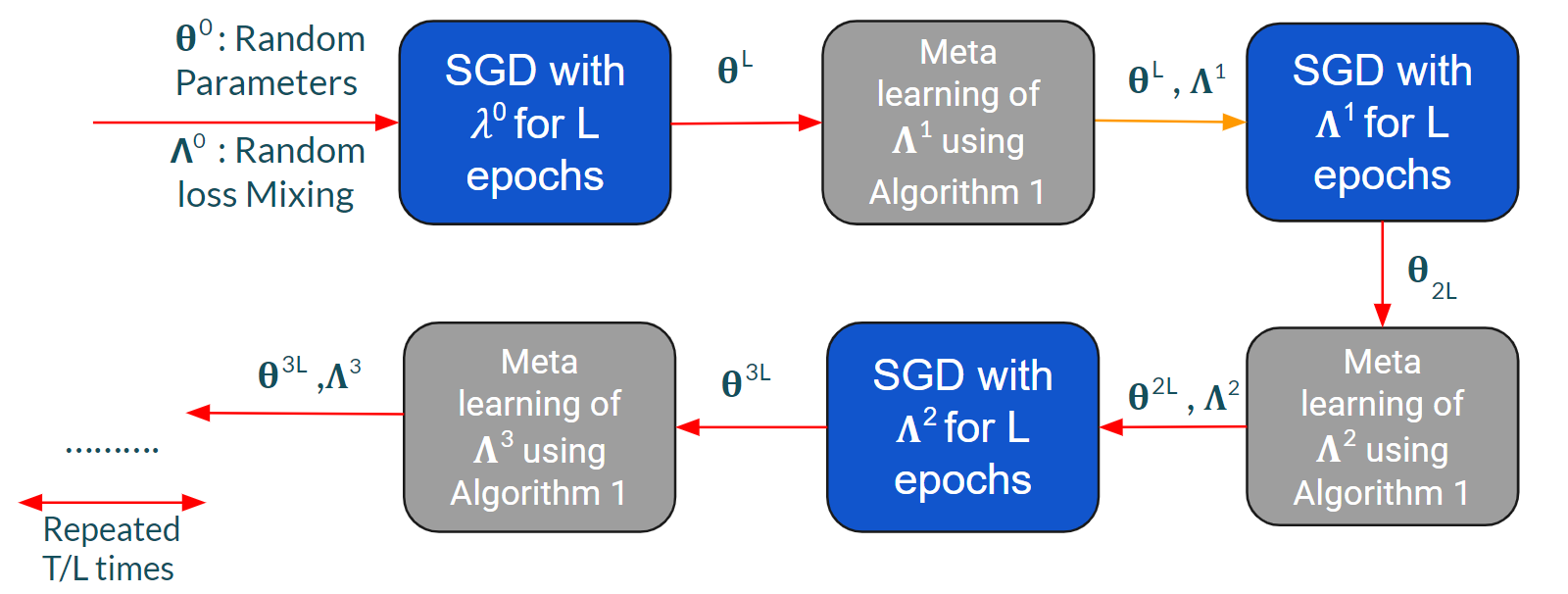}
  \caption{Flowchart of our approach to Adaptive Loss Mixing (\model)}
  \label{fig:flowchart}
\end{figure}

We expand Eq. \eqref{nested-equation} as follows, 
\begin{align}
    \small{\overbrace{\underset{{\lambda_p,\lambda_{a_1},\lambda_{a_2},\cdots,\lambda_{a_K}}}{\operatorname{argmin\hspace{0.7mm}}} \mathcal{L}_{CE}\big(\underbrace{\underset{\theta}{\operatorname{argmin\hspace{0.7mm}}} \mathcal{L}( \theta(\lambda_p,\lambda_{a_1},\lambda_{a_2},\cdots,\lambda_{a_K}))}_{inner-level}, {\mathcal V}\big)}^{outer-level}}
    \label{nested_equation_exp}
\end{align}

Then following the iterative approach illustrated in the Figure \ref{fig:flowchart} we have a one step update on validation as,  

\begin{align}
      &\hat{\theta^{t}}\left(\lambda_p^{\lfloor\frac{t}{L}\rfloor},\lambda_{a_1}^{\lfloor\frac{t}{L}\rfloor},\lambda_{a_2}^{\lfloor\frac{t}{L}\rfloor}, \cdots, \lambda_{a_K}^{\lfloor\frac{t}{L}\rfloor} \right) = \theta^{t} - \frac{\eta}{n}\sum_{i=1}^{n} \nabla_{\theta^{t}} \mathcal L_{i}(\theta^{t}(\lambda_{p_i}^{\lfloor\frac{t}{L}\rfloor},\lambda_{a_{1,i}}^{\lfloor\frac{t}{L}\rfloor},\lambda_{a_{2,i}}^{\lfloor\frac{t}{L}\rfloor}, \cdots, \lambda_{a_{K,i}}^{\lfloor\frac{t}{L}\rfloor})) \nonumber
\end{align}

Using the approximate model parameters obtained using the one step look-ahead SGD update, the outer optimization problem is solved as, for the parameters associated with the primary loss,
{\small
\begin{align}
  \nabla_{\lambda_{pi}^{\lfloor\frac{t}{L}\rfloor}}\mathcal L_{CE}&(\hat{\theta^{t}},{\mathcal V})  = \nabla_{\hat{\theta^{t}}}\mathcal L_{CE}(\hat{\theta^{t}},{\mathcal V}). \nabla_{\lambda_{pi}^{\lfloor\frac{t}{L}\rfloor}}\hat{\theta^{t}} = - \frac{\eta}{n}.\nabla_{\hat{\theta^{t}}}\mathcal L_{CE}(\hat{\theta^{t}},{\mathcal V}). \nabla_{\lambda_{pi}^{\lfloor\frac{t}{L}\rfloor}} \nabla_{\theta^{t}}\mathcal L_{i}^T
   \nonumber
\end{align}
}
This can be re-written as, 
\begin{align}
    \label{eqdis6}
      &\nabla_{\lambda_{pi}^{\lfloor\frac{t}{L}\rfloor}}\mathcal L_{CE}(\hat{\theta^{t}},{\mathcal V}) = - \frac{\eta}{n}.\nabla_{\hat{\theta^{t}}}\mathcal L_{CE}(\hat{\theta^{t}},{\mathcal V}).  [\nabla_{\lambda_{pi}^{\lfloor\frac{t}{L}\rfloor}} . (\lambda_{pi}^{\lfloor\frac{t}{L}\rfloor} \nabla_{\theta^{t}} \mathcal L_{pi})]^T 
\end{align}
and for parameters associated with $k^{th}$ auxiliary loss as, 
\begin{align}
      \label{eqdis7}
      &\nabla_{\lambda_{a_{k,i}}^{\lfloor\frac{t}{L}\rfloor}}\mathcal L_{CE}(\hat{\theta^{t}},{\mathcal V}) = - \frac{\eta}{n}.\nabla_{\hat{\theta^{t}}}\mathcal L_{CE}(\hat{\theta^{t}},{\mathcal V}).  [\nabla_{\lambda_{a_{k,i}}^{\lfloor\frac{t}{L}\rfloor}} . (\lambda_{a_{k,i}}^{\lfloor\frac{t}{L}\rfloor}\nabla_{\theta^{t}} \mathcal L_{a_{k,i}})]^T 
\end{align}
Here we perform a meta learning update to obtain the optimal $\lambda_p$ and $\lambda_a$ values to solve the outer optimisation problem. Using the meta-gradient (in Eq.\eqref{eqdis6} \& Eq.\eqref{eqdis7}), 
we update the $\lambda$s  for each of the training samples using the first order gradient update rule (see Eq.\eqref{eqdis4p} \& Eq.\eqref{eqdis4}). Here, $\eta_{\lambda}$ is the learning rate for smoothing parameters across all classes.
\begin{align}
      \lambda_{pi}^{\lfloor\frac{t}{L}\rfloor+1} = \lambda_{pi}^{\lfloor\frac{t}{L}\rfloor} - \eta_{\lambda} \nabla_{\lambda_{pi}^{\lfloor\frac{t}{L}\rfloor}} \mathcal L_{CE}(\hat{\theta^{t}},{\mathcal V}) 
    \label{eqdis4p}
\end{align}
\begin{align}
    \label{eqdis4}
      \lambda_{a_{k,i}}^{\lfloor\frac{t}{L}\rfloor+1} = \lambda_{a_{k,i}}^{\lfloor\frac{t}{L}\rfloor} - \eta_{\lambda} \nabla_{\lambda_{a_{k,i}}^{\lfloor\frac{t}{L}\rfloor}} \mathcal L_{CE}(\hat{\theta^{t}},{\mathcal V})
\end{align}
We update $\lambda$ values every L epochs. The updated  $\lambda_i^{\lfloor\frac{t}{L}\rfloor+1}$ values are then used to update the model parameters as shown in Eq.\eqref{eqdis5}. 
\begin{align}
    \label{eqdis5}
     \theta^{t+1} &= \theta^{t} - \frac{\eta}{n}\sum_{i=1}^{n} \nabla_{\theta^{t}}  \mathcal L_{i}(\theta^{t}(\lambda_{pi}^{\lfloor\frac{t}{L}\rfloor},,\lambda_{a_{1,i}}^{\lfloor\frac{t}{L}\rfloor},\lambda_{a_{2,i}}^{\lfloor\frac{t}{L}\rfloor}, \cdots, \lambda_{a_{K,i}}^{\lfloor\frac{t}{L} \rfloor}))\nonumber
\end{align}

\section{Convergence}\label{convergence}

\label{sec:lemproof}
In this section we lay out the complete proof of theorems. Our proof technique is inspired partly from the prior literature \cite{meta_net}. 


\begin{lemma}\label{lem:convergence}
Let validation loss be defined as 
\begin{align}
     \Lcal_{\lambda}(\hat{\theta^{t}}(\lambda_{pi}^{\lfloor\frac{t}{L}\rfloor},\lambda_{a_{1,i}}^{\lfloor\frac{t}{L}\rfloor},\lambda_{a_{2,i}}^{\lfloor\frac{t}{L}\rfloor}, \cdots, \lambda_{a_{K,i}}^{\lfloor\frac{t}{L} \rfloor})) = \frac{1}{V}\sum_i^V \Lcal_{CE}(\hat{\theta^{t}}(\lambda_{pi}^{\lfloor\frac{t}{L}\rfloor},\lambda_{a_{1,i}}^{\lfloor\frac{t}{L}\rfloor},\lambda_{a_{2,i}}^{\lfloor\frac{t}{L}\rfloor}, \cdots, \lambda_{a_{K,i}}^{\lfloor\frac{t}{L} \rfloor}),x_i,y_i)
\end{align}
 where ${(x_i,y_i)}_{i=1}^V$ constitute the validation set $\Vcal$. Suppose the validation loss function is Lipschitz-smooth with constant $\mu$, and the gradient associated with the train/validation loss function $\Lcal_i,L_{\lambda}$  have $\sigma$-bounded gradients with respect to training/validation data $x_i$. Then the gradient of $\Theta$ with respect to $\Lcal_V$ is Lipschitz continuous.

\end{lemma}

\begin{proof}

From Eq.s \eqref{eqdis4p} and \eqref{eqdis4} we know gradient of meta loss can be written as, 
\begin{align}
     \nabla_{\lambda_{pi}^{\lfloor\frac{t}{L}\rfloor}}\Lcal_{\lambda}(\hat{\theta^{t}}(\lambda_{pi}^{\lfloor\frac{t}{L}\rfloor},\lambda_{a_{1,i}}^{\lfloor\frac{t}{L}\rfloor},\lambda_{a_{2,i}}^{\lfloor\frac{t}{L}\rfloor}, \cdots, \lambda_{a_{K,i}}^{\lfloor\frac{t}{L} \rfloor})) & =  \nabla_{\lambda_{pi}^{\lfloor\frac{t}{L}\rfloor}} \mathcal L_{CE}(\hat{\theta^{t}},{\mathcal V}))*\lambda_{pi}^{\lfloor\frac{t}{L}\rfloor}* \nabla_{\theta^{t}}  \mathcal L_{pi}(\theta^{t}) 
     \\
     \nabla_{\lambda_{a_{k,i}}^{\lfloor\frac{t}{L}\rfloor}} \Lcal_{\lambda}(\hat{\theta^{t}}(\lambda_{pi}^{\lfloor\frac{t}{L}\rfloor},\lambda_{a_{1,i}}^{\lfloor\frac{t}{L}\rfloor},\lambda_{a_{2,i}}^{\lfloor\frac{t}{L}\rfloor}, \cdots, \lambda_{a_{K,i}}^{\lfloor\frac{t}{L} \rfloor})) & = \nabla_{\lambda_{a_{k,i}}^{\lfloor\frac{t}{L}\rfloor}} \mathcal L_{CE}(\hat{\theta^{t}},{\mathcal V})) * \lambda_{a_{k,i}}^{\lfloor\frac{t}{L}\rfloor}* \nabla_{\theta^{t}}  \Lcal_{a_{k,i}}(\theta^{t}) 
\end{align}



We show the proof only for $\lambda_{pi}^{\lfloor\frac{t}{L}\rfloor}$ , as for  $\lambda_{a_{k,i}}^{\lfloor\frac{t}{L}\rfloor}$ could be derived similarly. Also without loss of genrality, let $\lambda = \lambda_{pi}^{\lfloor\frac{t}{L}\rfloor}$ and $\Lcal_{\lambda}(\hat{\theta^{t}}(\lambda_{pi}^{\lfloor\frac{t}{L}\rfloor},\lambda_{a_{1,i}}^{\lfloor\frac{t}{L}\rfloor},\lambda_{a_{2,i}}^{\lfloor\frac{t}{L}\rfloor}, \cdots, \lambda_{a_{K,i}}^{\lfloor\frac{t}{L} \rfloor})) = \Lcal_{\lambda}(\hat{\theta^{t}}(\Lambda))$. Then, 

\begin{align}
     \nabla_{\lambda}\Lcal_{\lambda}(\hat{\theta^{t}}(\Lambda)) & =  \nabla_{\hat{\theta^{t}}} \mathcal L_{CE}(\hat{\theta^{t}},{\mathcal V}))* \nabla_{\theta^{t}}  \mathcal L_{pi}(\theta^{t}) 
\end{align}

Taking gradient of $\lambda$ on both sides, we have

\begin{align}
     \nabla^2_{\lambda_2}\Lcal_{\lambda}(\hat{\theta^{t}}(\Lambda)) & =  \nabla_{\hat{\theta^{t}}} \mathcal L_{CE}(\hat{\theta^{t}},{\mathcal V}))* (\nabla_{\theta^{t}} \mathcal L_{pi}(\theta^{t}) )^2
\end{align}

Since, $\|\nabla_{\hat{\theta^{t}}} \mathcal L_{CE}(\hat{\theta^{t}},{\mathcal V}))\| \leq \sigma$  and $\|(\nabla_{\theta^{t}} \mathcal L_{pi}(\theta^{t}) \| \leq \sigma$

\begin{align}
     \|\nabla^2_{\lambda_2}\Lcal_{\lambda}(\hat{\theta^{t}}(\Lambda))\| \leq \sigma^3
\end{align}

Let $L_{V} = \sigma^3$, then based on Lagrange mean value theorem, we have:

\begin{align}
\|\nabla_{\lambda}\Lcal_{\lambda}(\hat{\theta^{t}}(\Lambda_1))-\nabla_{\lambda^2}\Lcal_{\lambda}(\hat{\theta^{t}}(\Lambda_2))\| \leq L_V \|\Lambda_1-\Lambda_2\|, \ \ for \ all \ \ \Lambda_1, \Lambda_2,
\end{align}
where $\nabla_{\lambda}\Lcal_{\lambda}(\hat{\theta^{t}}(\Lambda_1))=  \nabla_{\lambda}\Lcal_{\lambda}(\hat{\theta^{t}}(\Lambda))\big|_{\Lambda_1}$.
\end{proof}

\begin{theorem} \label{thm::convergence}
		Suppose the validation loss function is Lipschitz-smooth with constant $\mu$, and the gradient associated with the train/validation loss function $\Lcal_i,L_{\lambda}$  have $\sigma$-bounded gradients with respect to training/validation data $x_i$. Let the learning rate $\eta$ satisfies $\eta=\min\{1,\frac{k}{T}\}$, for some $k>0$, such that $\frac{k}{T}<1$, and $\eta_{\lambda}, 1\leq t\leq N$ is a monotone descent sequence, $\eta_{\lambda} =\min\{\frac{1}{\mu},\frac{c}{\sigma\sqrt{T}}\} $ for some $c>0$, such that $\frac{\sigma\sqrt{T}}{c}\geq \mu$ and $\sum_{t=1}^\infty \eta_{\lambda} \leq \infty,\sum_{t=1}^\infty \eta_{\lambda}^2 \leq \infty $. Then \model{} can achieve $\mathbb{E}[ \|\nabla L_{CE}(\hat{\theta}^{t}(\Lambda^{t}),{\mathcal V})\|_2^2] \leq \epsilon$ in $\mathcal{O}(1/\epsilon^2)$ steps. More specifically,
		\begin{align}
		\min_{0\leq t \leq T} \mathbb{E}[ \|\nabla L_{CE}(\hat{\theta}^{t}(\Lambda^{t}),{\mathcal V})\|_2^2] \leq \mathcal{O}(\frac{C}{\sqrt{T}}),
		\end{align}
		where $C$ is some constant independent of the convergence process, $\delta$ is the variance of drawing uniformly mini-batch sample at random.
\end{theorem}

\begin{proof}
		The update of $\Lambda$ in each iteration is as follows:
		\begin{align}
          \Lambda_i^{\lfloor\frac{t}{L}\rfloor +1 } = \Lambda_i^{\lfloor\frac{t}{L}\rfloor} - \eta_{\lambda} \nabla_{\Lambda_i^{\lfloor\frac{t}{L}\rfloor}} \mathcal L_{CE}(\hat{\theta^{t}},{\mathcal V}) 
        \end{align}
		
		This can be written as:
		\begin{align}
              \Lambda_i^{\lfloor\frac{t}{L}\rfloor +1 } = \Lambda_i^{\lfloor\frac{t}{L}\rfloor} - \eta_{\lambda} \nabla_{\Lambda_i^{\lfloor\frac{t}{L}\rfloor}} \mathcal L_{CE}(\hat{\theta^{t}},{\mathcal V})|_{\Xi_t}
        \end{align}
		
		Since the mini-batch $\Xi_t$ is drawn uniformly from the entire data set, we can rewrite the update equation as:
		\begin{align}
		\Lambda_i^{\lfloor\frac{t}{L}\rfloor +1 } = \Lambda_i^{\lfloor\frac{t}{L}\rfloor}  -\eta_{\lambda} [\nabla_{\Lambda_i^{\lfloor\frac{t}{L}\rfloor}} \mathcal L_{CE}(\hat{\theta^{t}},{\mathcal V})+\xi^{(t)}],
		\end{align}
		where $\xi^{(t)} =  \nabla_{\Lambda_i^{\lfloor\frac{t}{L}\rfloor}} \mathcal L_{CE}(\hat{\theta^{t}},{\mathcal V})\big|_{\Xi_t} -\nabla_{\Lambda_i^{\lfloor\frac{t}{L}\rfloor}} \mathcal L_{CE}(\hat{\theta^{t}},{\mathcal V})$. Note that $\xi^{(t)}$ are i.i.d random variable with finite variance, since $\Xi_t$ are drawn i.i.d with a finite number of samples. Furthermore, $\mathbb{E}[\xi^{(t)}]=0$, since samples are drawn uniformly at random.
		Observe that
		
		\begin{align}\label{eq:add_sub}
		\begin{split}
		L_{CE}(\hat{\theta}^{t+1}(\Lambda^{\lfloor\frac{t}{L}\rfloor+1}),{\mathcal V})-L_{CE}(\hat{\theta^{t}}(\Lambda^{\lfloor\frac{t}{L}\rfloor}),{\mathcal V})
		&= \left\{L_{CE}(\hat{\theta}^{t+1}(\Lambda^{\lfloor\frac{t}{L}\rfloor+1}),{\mathcal V})- L_{CE}(\hat{\theta^{t}}(\Lambda^{\lfloor\frac{t}{L}\rfloor+1}),{\mathcal V})\right\}
		\\&+\left\{L_{CE}(\hat{\theta^{t}}(\Lambda^{\lfloor\frac{t}{L}\rfloor+1}),{\mathcal V})-L_{CE}(\hat{\theta^{t}}(\Lambda^{\lfloor\frac{t}{L}\rfloor}),{\mathcal V})\right\}.
		\end{split}
		\end{align}
		
		By Lipschitz smoothness of validation loss function,  we have
		
		\begin{align*}
		L_{CE}(\hat{\theta}^{t+1}(\Lambda^{\lfloor\frac{t}{L}\rfloor+1}),{\mathcal V})- L_{CE}(\hat{\theta^{t}}(\Lambda^{\lfloor\frac{t}{L}\rfloor+1}),{\mathcal V})
		&\leq \langle \nabla L_{CE}(\hat{\theta^{t}}(\Lambda^{\lfloor\frac{t}{L}\rfloor+1}),{\mathcal V}), \hat{\theta}^{t+1}(\Lambda^{\lfloor\frac{t}{L}\rfloor+1})-\hat{\theta}^{t}(\Lambda^{\lfloor\frac{t}{L}\rfloor+1}) \rangle\\&+ \frac{\mu}{2}\|\hat{\theta}^{t+1}(\Lambda^{\lfloor\frac{t}{L}\rfloor+1})-\hat{\theta}^{t}(\Lambda^{\lfloor\frac{t}{L}\rfloor+1})\|_2^2 
		\end{align*}
		
		Since  $\hat{\theta}^{t+1}(\Lambda^{\lfloor\frac{t}{L}\rfloor+1})-\hat{\theta}^{t}(\Lambda^{\lfloor\frac{t}{L}\rfloor+1})= - \frac{\eta}{n}\sum_{i=1}^{n} \nabla_{\theta^{t}}  \mathcal L_{i}(\theta^{t},\Lambda_{i}^{\lfloor\frac{t}{L}\rfloor}) $  according to Eq.(\ref{eqdis2}),(\ref{eq:final}), we have
		\begin{align}
		\|L_{CE}(\hat{\theta}^{t+1}(\Lambda^{\lfloor\frac{t}{L}\rfloor+1}),{\mathcal V})- L_{CE}(\hat{\theta^{t}}(\Lambda^{\lfloor\frac{t}{L}\rfloor+1}),{\mathcal V}) \|\leq \eta \sigma^2+ \frac{\mu\eta^2}{2} \sigma^2 = \eta\sigma^2 (1+\frac{\eta \mu}{2})
		\end{align}
		due to the bound on gradients
		
		By Lipschitz continuity of $\nabla L_{CE}(\hat{\theta^{t}}(\Lambda^{\lfloor\frac{t}{L}\rfloor+1}),{\mathcal V})$ according to Lemma \ref{lem:convergence}, we can obtain the following:
		
		\begin{align*}
		&L_{CE}(\hat{\theta}^{t+1}(\Lambda^{\lfloor\frac{t}{L}\rfloor+1}),{\mathcal V})- L_{CE}(\hat{\theta^{t}}(\Lambda^{\lfloor\frac{t}{L}\rfloor+1}),{\mathcal V})  \leq \langle\nabla L_{CE}(\hat{\theta^{t}}(\Lambda^{\lfloor\frac{t}{L}\rfloor+1}),{\mathcal V}),\Lambda^{\lfloor\frac{t}{L}\rfloor+1}-\Lambda^{\lfloor\frac{t}{L}\rfloor} \rangle + \frac{\mu}{2} \|\Lambda^{\lfloor\frac{t}{L}\rfloor+1}-\Lambda^{\lfloor\frac{t}{L}\rfloor}\|_2^2\\
		& = \langle\nabla L_{CE}(\hat{\theta^{t}}(\Lambda^{\lfloor\frac{t}{L}\rfloor+1}),{\mathcal V}),  - \eta_{\lambda} \nabla_{\Lambda_i^{\lfloor\frac{t}{L}\rfloor}} \mathcal L_{CE}(\hat{\theta^{t}},{\mathcal V})+\xi^{(t)} ] \rangle + \frac{\mu\eta_{\lambda}^2}{2} \|  \nabla_{\Lambda_i^{\lfloor\frac{t}{L}\rfloor}} \mathcal L_{CE}(\hat{\theta^{t}},{\mathcal V})+\xi^{(t)}\|_2^2\\
		& = -(\eta_{\lambda}^2-\frac{\mu\eta_{\lambda}^2}{2}) \|\nabla L_{CE}(\hat{\theta^{t}}(\Lambda^{\lfloor\frac{t}{L}\rfloor+1}),{\mathcal V})\|_2^2 + \frac{\mu\eta_{\lambda}^2}{2}\|\xi^{(t)}\|_2^2 - (\eta_{\lambda}-\mu\eta_{\lambda}^2)\langle \nabla L_{CE}(\hat{\theta^{t}}(\Lambda^{\lfloor\frac{t}{L}\rfloor+1}),{\mathcal V}),\xi^{(t)}\rangle.
		\end{align*}
		
		Thus eq. \ref{eq:add_sub} can be written as 
		
		\begin{align}\label{eq:add_sub}
		\begin{split}
		L_{CE}(\hat{\theta}^{t+1}(\Lambda^{\lfloor\frac{t}{L}\rfloor+1}),{\mathcal V})-L_{CE}(\hat{\theta^{t}}(\Lambda^{\lfloor\frac{t}{L}\rfloor}),{\mathcal V})
		&\leq \eta\sigma^2 (1+\frac{\eta \mu}{2}) -(\eta_{\lambda}^2-\frac{\mu\eta_{\lambda}^2}{2}) \|\nabla L_{CE}(\hat{\theta^{t}}(\Lambda^{\lfloor\frac{t}{L}\rfloor+1}),{\mathcal V})\|_2^2 \\ &+ \frac{\mu\eta_{\lambda}^2}{2}\|\xi^{(t)}\|_2^2 - (\eta_{\lambda}-\mu\eta_{\lambda}^2)\langle \nabla L_{CE}(\hat{\theta^{t}}(\Lambda^{\lfloor\frac{t}{L}\rfloor+1}),{\mathcal V}),\xi^{(t)}\rangle.
		\end{split}
		\end{align}
		
		Rearranging the terms, we can obtain

		\begin{align}\label{eq:add_sub}
		\begin{split}
        (\eta_{\lambda}^2-\frac{\mu\eta_{\lambda}^2}{2}) \|\nabla L_{CE}(\hat{\theta^{t}}(\Lambda^{\lfloor\frac{t}{L}\rfloor+1}),{\mathcal V})\|_2^2
		&\leq \eta\sigma^2 (1+\frac{\eta \mu}{2}) -L_{CE}(\hat{\theta}^{t+1}(\Lambda^{\lfloor\frac{t}{L}\rfloor+1}),{\mathcal V})+L_{CE}(\hat{\theta^{t}}(\Lambda^{\lfloor\frac{t}{L}\rfloor}),{\mathcal V})  \\ &+ \frac{\mu\eta_{\lambda}^2}{2}\|\xi^{(t)}\|_2^2 - (\eta_{\lambda}-\mu\eta_{\lambda}^2)\langle \nabla L_{CE}(\hat{\theta^{t}}(\Lambda^{\lfloor\frac{t}{L}\rfloor+1}),{\mathcal V}),\xi^{(t)}\rangle.
		\end{split}
		\end{align}
		
		Summing up the above inequalities and rearranging the terms, we can obtain
		
		\begin{align}\label{eqrand}
		\begin{split}
		&\sum\nolimits_{t=1}^T(\eta_{\lambda}^2-\frac{\mu\eta_{\lambda}^2}{2}) \|\nabla L_{CE}(\hat{\theta^{t}}(\Lambda^{t}),{\mathcal V})\|_2^2
		\leq L_{CE}(\hat{\theta}^{1}(\Lambda^{1}),{\mathcal V})- L_{CE}(\hat{\theta}^{T+1}(\Lambda^{T+1}),{\mathcal V})\\
		&+\sum_{t=1}^T\eta\sigma^2 (1+\frac{\eta \mu}{2})-\!\sum_{t=1}^T\!(\eta_{\lambda}\!-\!\mu\eta_{\lambda}^2)\!\langle\nabla\! L_{CE}(\hat{\theta}^{t}(\Lambda^{t}),{\mathcal V}),\!\xi^{(t)}\!\rangle\!+\!\frac{\mu}{2}\!\sum_{t=1}^T\!\eta_{\lambda}^2\|\xi\!^{(t)}\!\|_2^2 \\
		&\leq L_{CE}(\hat{\theta}^{1}(\Lambda^{1}),{\mathcal V})\!+\sum_{t=1}^T\eta\sigma^2 (1+\frac{\eta \mu}{2})-\!\sum_{t=1}^T \!(\eta_{\lambda}\!-\!\mu\eta_{\lambda}^2)\!\langle\!\nabla\! L_{CE}(\hat{\theta}^{t}(\Lambda^{t}),{\mathcal V}),\!\xi^{(t)}\!\rangle\!+\!\frac{\mu}{2}\!\sum_{t=1}^T\!\eta_{\lambda}^2\|\xi^{(t)}\|_2^2,
		\end{split}
		\end{align}
		
		Taking expectations with respect to $\xi^{(N)}$ on both sides of Eq. \ref{eqrand}, we can then obtain:
		
		\begin{align}
		\sum_{t=1}^T (\eta_{\lambda}-\frac{\mu\eta_{\lambda}^2}{2})\mathbb{E}_{\xi^{(N)}} \|\nabla L_{CE}(\hat{\theta}^{t}(\Lambda^{t}),{\mathcal V})\|_2^2 \leq L_{CE}(\hat{\theta}^{1}(\Lambda^{1}),{\mathcal V})+\sum_{t=1}^T\eta\sigma^2 (1+\frac{\eta \mu}{2}) + \frac{L\sigma^2}{2} \sum_{t=1}^T \eta_{\lambda}^2,
		\end{align}
		
		since $\mathbb{E}_{\xi^{(N)}} \langle \nabla \mathcal{L}^{meta}(\Theta^{(t)}),\xi^{(t)}\rangle =0 $ and $\mathbb{E} [\|\xi^{(t)}\|_2^2] \leq \delta^2$, where
		$\delta^2$ is the variance of $\xi^{(t)}$.
		Furthermore, we can deduce that
		\begin{align}
		\begin{split}
		\min_{t}&\quad \mathbb{E} [ \|\nabla L_{CE}(\hat{\theta}^{t}(\Lambda^{t}),{\mathcal V})\|_2^2] \leq \frac{\sum_{t=1}^T (\eta_{\lambda}-\frac{\mu\eta_{\lambda}^2}{2})\mathbb{E}_{\xi^{(N)}} \|\nabla L_{CE}(\hat{\theta}^{t}(\Lambda^{t}),{\mathcal V})\|_2^2}{\sum_{t=1}^T (\eta_{\lambda}-\frac{\mu\eta_{\lambda}^2}{2})}\\
		&\leq  \frac{1}{\sum_{t=1}^T (2\eta_{\lambda}-\mu\eta_{\lambda}^2)} \left[2L_{CE}(\hat{\theta}^{1}(\Lambda^{1}),{\mathcal V})+\sum_{t=1}^T\eta\sigma^2 (2+\eta \mu\!) + \mu\delta^2\sum_{t=1}^T \eta_{\lambda}^2\right]\\
		& \leq \frac{1}{\sum_{t=1}^T \eta_{\lambda}} \left[2L_{CE}(\hat{\theta}^{1}(\Lambda^{1}),{\mathcal V})+\sum_{t=1}^T\eta\sigma^2 (2+\eta \mu) + \mu\delta^2\sum_{t=1}^T \eta_{\lambda}^2\right] \\
		& \leq \frac{1}{T\eta_{\lambda}} \left[2L_{CE}(\hat{\theta}^{1}(\Lambda^{1}),{\mathcal V})+ \eta \sigma^2 T (2+ \mu) + \mu\delta^2 \sum_{t=1}^T \eta_{\lambda}^2\right]\\
		& = \frac{2L_{CE}(\hat{\theta}^{1}(\Lambda^{1}),{\mathcal V})}{T} \frac{1}{\eta_{\lambda}} + \frac{2\eta \sigma^2 (2+\mu)}{\eta_{\lambda}}+\frac{\mu\delta^2}{T}\sum_{t=1}^T \eta_{\lambda}\\
		& \leq  \frac{2L_{CE}(\hat{\theta}^{1}(\Lambda^{1}),{\mathcal V})}{T} \frac{1}{\eta_{\lambda}} + \frac{2\eta \sigma^2 (2+\mu)}{\eta_{\lambda}}+\mu\delta^2 \eta_{\lambda}\\
		& = \frac{L_{CE}(\hat{\theta}^{1}(\Lambda^{1}),{\mathcal V})}{T} \max\{\mu,\frac{\delta\sqrt{T}}{c}\} + \min\{1,\frac{k}{T}\} \max\{\mu,\frac{\delta\sqrt{T}}{c}\} \sigma^2  (2+\mu)+ \mu \delta^2 \min\{\frac{1}{mu},\frac{c}{\sigma\sqrt{T}}\}\\
		& \leq \frac{\delta\mathcal{L}^{meta}\!(\hat{\mathbf{w}}^{(1)}(\!\Theta^{(1)}\!)\!}{c\sqrt{T}}+ \frac{k\delta \sigma^2(2+\mu)}{c \sqrt{T}} + \frac{\mu\delta c}{\sqrt{T}} = \mathcal{O}(\frac{1}{\sqrt{T}}).
		\end{split}
		\end{align}
		The third inequlity holds for $\sum_{t=1}^T (2\eta_{\lambda}-\mu\eta_{\lambda}^2) \geq \sum_{t=1}^T \eta_{\lambda}$.
		Therefore, we can conclude that our algorithm can always achieve $\min_{0\leq t \leq T} \mathbb{E}[ \|L_{CE}(\hat{\theta}^{t}(\Lambda^{t}),{\mathcal V})\|_2^2] \leq \mathcal{O}(\frac{1}{\sqrt{T}})$ in $T$ steps.
		
\end{proof} 

\begin{lemma}\label{lemma2}
		(Lemma A.5 in \cite{mairal2013stochastic}) Let $(a_n)_{n\leq 1},(b_n)_{n\leq 1}$ be two non-negative real sequences such that the series $\sum_{i=1}^{\infty} a_n$ diverges, the series $\sum_{i=1}^{\infty} a_nb_n$ converges, and there exists $K>0$ such that $|b_{n+1}-b_n|\leq K a_n$. Then the sequences $(b_n)_{n\leq 1}$ converges to 0.\\
	\end{lemma}
	\begin{theorem}\label{th2}
		Suppose the training loss function $\mathcal{L}$ is Lipschitz-smooth with constant $\mu$, and the gradient associated with the train/validation loss function $\Lcal_i,L_V$  have $\sigma$-bounded gradients with respect to training/validation data $x_i$.Let and $\mathcal{V}(\cdot)$ is differential with a $\delta$-bounded gradient. Let the learning rate $\eta$ satisfies $\eta=\min\{1,\frac{k}{T}\}$, for some $k>0$, such that $\frac{k}{T}<1$, and $\eta_{\lambda}, 1\leq t\leq N$ is a monotone descent sequence, $\eta_{\lambda} =\min\{\frac{1}{\mu},\frac{c}{\sigma\sqrt{T}}\} $ for some $c>0$, such that $\frac{\sigma\sqrt{T}}{c}\geq \mu$ and $\sum_{t=1}^\infty \eta_{\lambda} \leq \infty,\sum_{t=1}^\infty \eta_{\lambda}^2 \leq \infty $.Then
		\begin{align}
		\lim_{t\rightarrow \infty} \mathbb{E}[\| \nabla \mathcal{L}(\hat{\theta}^{t}(\Lambda^{t+1}))  \|_2^2]=0.
		\end{align}
	\end{theorem}
	
\begin{proof}

    We update model parameters as, 

    \begin{align}
     \theta^{t+1} = \theta^{t} - \frac{\eta}{n}\sum_{i=1}^{n} \nabla_{\theta^{t}}  \mathcal L_{i}(\theta^{t}(\Lambda_{i}^{\lfloor\frac{t}{L}\rfloor+1}))
    \end{align}
    
    It can be written as:
		\begin{align}
		\theta^{t+1} = \theta^{t} - \eta \nabla \mathcal L(\theta^{t}(\Lambda^{\lfloor\frac{t}{L}\rfloor+1}))|_{\Psi_t}
		\end{align}
		where $\nabla \mathcal L(\theta^{t},\Lambda^{\lfloor\frac{t}{L}\rfloor+1}) = \frac{1}{n}\sum_{i=1}^{n} \nabla_{\theta^{t}}  \mathcal L_{i}(\theta^{t},\Lambda_{i}^{\lfloor\frac{t}{L}\rfloor+1})$. Since the mini-batch $\Psi_t$ is drawn uniformly at random, we can rewrite the update equation as:
		\begin{align}
		\theta^{t+1} = \theta^{t} - \eta [ \nabla \mathcal L(\theta^{t}(\Lambda^{\lfloor\frac{t}{L}\rfloor+1}))+\psi^{(t)}],
		\end{align}
		where $\psi^{(t)} = \nabla \mathcal L(\theta^{t}(\Lambda^{\lfloor\frac{t}{L}\rfloor+1}))|_{\Psi_t}|_{\Psi_t}-\nabla \mathcal L(\theta^{t}(\Lambda^{\lfloor\frac{t}{L}\rfloor+1}))|_{\Psi_t}$.
		Note that $\psi^{(t)}$ is i.i.d. random variable with finite variance, since $\Psi_t$ are drawn i.i.d. with a finite number of samples. Furthermore, $\mathbb{E}[\psi^{(t)}]=0$, since samples are drawn uniformly at random, and $\mathbb{E}[\|\psi^{(t)}\|_2^2]\leq \rho^2$.
		
		The inner optimization $L(\theta(\Lambda))$ defined in Eq. \ref{nested_equation_exp} can be easily checked to be Lipschitz-smooth with constant $L$, and have $\sigma$-bounded gradients with respect to training data.
		Observe that
			\begin{align}
			\begin{split}
			&\mathcal L(\theta^{t+1}(\Lambda^{\lfloor\frac{t}{L}\rfloor+2})) -\mathcal L(\theta^{t}(\Lambda^{\lfloor\frac{t}{L}\rfloor+1})) = \left\{\mathcal L(\theta^{t+1}(\Lambda^{\lfloor\frac{t}{L}\rfloor+2}))-\mathcal L(\theta^{t+1}(\Lambda^{\lfloor\frac{t}{L}\rfloor+1}))\right\}
			+\left\{\mathcal L(\theta^{t+1}(\Lambda^{\lfloor\frac{t}{L}\rfloor+1}))-\mathcal L(\theta^{t}(\Lambda^{\lfloor\frac{t}{L}\rfloor+1}))\right\}.
			\end{split}
			\end{align}
		
	For the first term,
	\begin{align}
	\begin{split}
	\mathcal L(\theta^{t+1}(\Lambda^{\lfloor\frac{t}{L}\rfloor+2}))-\mathcal L(\theta^{t+1}(\Lambda^{\lfloor\frac{t}{L}\rfloor+1}))
	= & \frac{\eta_{\lambda}}{n}\sum_{i=1}^n \left\{\Lambda_i^{\lfloor\frac{t}{L}\rfloor+2}-\Lambda_i^{\lfloor\frac{t}{L}\rfloor+1}\right\} \mathcal{L}_i^{train}(\mathbf{w}^{(t)})\\
	= &\frac{1}{n}\sum_{i=1}^n \left\{\nabla_{\Lambda_{i}^{\lfloor\frac{t}{L}\rfloor}} \mathcal L_{CE}(\hat{\theta^{t}},{\mathcal V}) + \xi^{(t)}\right\} \mathcal{L}_i^{train}(\mathbf{w}^{(t)})
	\end{split}
	\end{align}
	
For the second term,
		\begin{align}
		\begin{split}
		\mathcal{L}(\theta^{t+1}(\Lambda^{\lfloor\frac{t}{L}\rfloor+1}))&-\mathcal {L}(\theta^{t}(\Lambda^{\lfloor\frac{t}{L}\rfloor+1})) \\
		&\leq \langle\nabla \mathcal{L}(\theta^{t}(\Lambda^{\lfloor\frac{t}{L}\rfloor+1})),\theta^{t+1}-\theta^{t} \rangle + \frac{\mu}{2} \|\theta^{(t+1)}-\theta^{(t)}\|_2^2\\
		& = \langle \nabla \mathcal{L}(\theta^{t}(\Lambda^{\lfloor\frac{t}{L}\rfloor+1})), -\eta [\nabla \mathcal{L}(\theta^{t}(\Lambda^{\lfloor\frac{t}{L}\rfloor+1}))+\psi^{(t)}] \rangle + \frac{\mu \eta^2}{2} \|\nabla \mathcal{L}(\theta^{t}(\Lambda^{\lfloor\frac{t}{L}\rfloor+1}))+\psi^{(t)}\|_2^2\\
		& = -(\eta-\frac{\mu\eta^2}{2}) \|\nabla \mathcal{L}(\theta^{t}(\Lambda^{\lfloor\frac{t}{L}\rfloor+1}))\|_2^2 + \frac{\mu\eta^2}{2}\|\psi^{(t)}\|_2^2 - (\eta-\mu\eta^2)\langle \nabla \mathcal{L}(\theta^{t}(\Lambda^{\lfloor\frac{t}{L}\rfloor+1})), \psi^{(t)}\rangle.
		\end{split}
		\end{align}
	
Therefore, we have:
	\begin{align}
	\begin{split}
	\mathcal L(\theta^{t+1}(\Lambda^{\lfloor\frac{t}{L}\rfloor+2})) - \mathcal L(\theta^{t}(\Lambda^{\lfloor\frac{t}{L}\rfloor+1}))
	& \leq  \frac{\eta_{\lambda}}{n}\sum_{i=1}^n \left\{\nabla_{\Lambda_{i}^{\lfloor\frac{t}{L}\rfloor}} \mathcal {L}_{CE}(\hat{\theta^{t}},{\mathcal V}) + \xi^{(t)}\right\} \mathcal{L}_i^{train}(\mathbf{w}^{(t)}) -(\eta-\frac{\mu \eta^2}{2}) \|\nabla \mathcal{L}(\theta^{t}(\Lambda^{\lfloor\frac{t}{L}\rfloor+1}))\|_2^2  \\
	&\frac{\mu\eta^2}{2}\|\psi^{(t)}\|_2^2 - (\eta- \mu \eta^2)\langle \nabla \mathcal{L}(\theta^{t}(\Lambda^{\lfloor\frac{t}{L}\rfloor+1})), \psi^{(t)}\rangle.
	\label{eq25}
	\end{split}
	\end{align}
	
Taking expectation of both sides of (\ref{eq25}) and since $\mathbb{E}[\xi^{(t)}]=0,\mathbb{E}[\psi^{(t)}]=0$, we have
	\begin{align*}
	\mathbb{E}[\mathcal L(\theta^{t+1}(\Lambda^{\lfloor\frac{t}{L}\rfloor+2}))]&-\mathbb{E}[\mathcal L(\theta^{t}(\Lambda^{\lfloor\frac{t}{L}\rfloor+1}))]
	\leq \mathbb{E} \bigg[ \frac{\eta_{\lambda}}{n}\sum_{i=1}^n \left\{\nabla_{\Lambda_{i}^{\lfloor\frac{t}{L}\rfloor}} \mathcal L_{CE}(\hat{\theta^{t}},{\mathcal V}) + \xi^{(t)}\right\} \mathcal{L}_i^{train}(\mathbf{w}^{(t)})\bigg]\\&  -\eta \mathbb{E}[\|\nabla \mathcal{L}(\theta^{t}(\Lambda^{\lfloor\frac{t}{L}\rfloor+1}))\|_2^2] + \frac{\mu \eta^2}{2} \left\{\mathbb{E}[\|\nabla \mathcal{L}(\theta^{t}(\Lambda^{\lfloor\frac{t}{L}\rfloor+1}))\|_2^2]+\mathbb{E}[ \|\psi^{(t)}\|_2^2 ]\right\}
	\end{align*}
	
Summing up the above inequalities over $t=1,...,\infty$ in both sides, we obtain 

\begin{align*}
\!\sum_{t=1}^{\infty} \!\!\eta \mathbb{E}[\|\nabla \mathcal{L}(\theta^{t}(\Lambda^{\lfloor\frac{t}{L}\rfloor+1}))\|_2^2]]   &\leq \sum_{t=1}^{\infty} \!\!\frac{\mu\eta^2}{2}\!\left\{ \mathbb{E}[\|\nabla\! \mathcal{L}(\theta^{t}(\Lambda^{\lfloor\frac{t}{L}\rfloor+1}))\!\|_2^2]\!+\!\mathbb{E}[\|\psi^{(t)}\|_2^2 ]\right\} \!+\\ &  \mathbb{E}\bigg[ \frac{\eta_{\lambda}}{n}\sum_{i=1}^n \left\{\|\nabla_{\Lambda_{i}^{\lfloor\frac{t}{L}\rfloor}} \mathcal L_{CE}(\hat{\theta^{t}},{\mathcal V})\| + \xi^{(t)}\right\} \|\mathcal{L}_i^{train}(\mathbf{w}^{(t)})\|\bigg]
\\ & \leq \sum_{t=1}^{\infty} \!\!\frac{\mu\eta^2}{2}\!\left\{ \sigma^2 +\rho^2]\right\} \!+ \sum_{t=1}^{\infty} \eta_{\lambda}\sigma^2 < \infty
\end{align*} 

The last inequality holds since $\sum_{t=0}^{\infty}\eta^2 < \infty$ and $\sum_{t=0}^{\infty}\eta_{\lambda} < \infty$. Thus we have 

\begin{align*}
\!\sum_{t=1}^{\infty} \!\!\eta \mathbb{E}[\|\nabla \mathcal{L}(\theta^{t}(\Lambda^{\lfloor\frac{t}{L}\rfloor+1}))\|_2^2]] < \infty
\end{align*} 

By Lemma \ref{lemma2}, to substantiate $\lim_{t\rightarrow \infty} \mathbb{E}[\|\nabla L(\theta^{t}(\Lambda^{\lfloor\frac{t}{L}\rfloor+1}))\|_2^2]]]=0$, since $\sum_{t=0}^{\infty}\eta = \infty$, it only needs to prove:

\begin{align}
\left|\mathbb{E}[\nabla L(\theta^{t+1}(\Lambda^{\lfloor\frac{t}{L}\rfloor+2}))\|_2^2]-\mathbb{E}[\nabla L(\theta^{t}(\Lambda^{\lfloor\frac{t}{L}\rfloor+1}))\|_2^2]  \right|\leq C \alpha_k,
\end{align}
for some constant $C$. Based on the inequality:
\begin{align}
\left|(\|a\|+\|b\|)(\|a\|-\|b\|)\right| \leq \|a+b\|\|a-b\|,
\end{align}

we then have:

\begin{align}
\begin{split}
&\left|\mathbb{E}[\nabla \mathcal{L} (\theta^{t+1}(\Lambda^{\lfloor\frac{t}{L}\rfloor+2}))\|_2^2]-\mathbb{E}[\nabla \mathcal{L}(\theta^{t}(\Lambda^{\lfloor\frac{t}{L}\rfloor+1}))\|_2^2]   \right|\\
&= \left|\mathbb{E}\left[( \|\nabla \mathcal{L}(\theta^{t+1}(\Lambda^{\lfloor\frac{t}{L}\rfloor+2}))\|_2 + \|\nabla \mathcal{L}(\theta^{t}(\Lambda^{\lfloor\frac{t}{L}\rfloor+1}))\|_2) ( \|\nabla \mathcal{L}(\theta^{t+1}(\Lambda^{\lfloor\frac{t}{L}\rfloor+2}))\|_2 - \|\nabla \mathcal{L}(\theta^{t}(\Lambda^{\lfloor\frac{t}{L}\rfloor+1}))\|_2)\right]\right|\\
& \leq \mathbb{E}\left[ \left|\|\nabla \mathcal{L}(\theta^{t+1}(\Lambda^{\lfloor\frac{t}{L}\rfloor+2}))\|_2 + \|\nabla \mathcal{L}(\theta^{t}(\Lambda^{\lfloor\frac{t}{L}\rfloor+1}))\|_2)\right|\left| ( \|\nabla \mathcal{L}(\theta^{t+1}(\Lambda^{\lfloor\frac{t}{L}\rfloor+2}))\|_2 - \|\nabla \mathcal{L}(\theta^{t}(\Lambda^{\lfloor\frac{t}{L}\rfloor+1}))\|_2)\right|\right]\\
& \leq \mathbb{E}\left[ \left\|\|\nabla \mathcal{L}(\theta^{t+1}(\Lambda^{\lfloor\frac{t}{L}\rfloor+2}))\|_2 + \|\nabla \mathcal{L}(\theta^{t}(\Lambda^{\lfloor\frac{t}{L}\rfloor+1}))\|_2)\right\|_2 \left\| ( \|\nabla \mathcal{L}(\theta^{t+1}(\Lambda^{\lfloor\frac{t}{L}\rfloor+2}))\|_2 - \|\nabla \mathcal{L}(\theta^{t}(\Lambda^{\lfloor\frac{t}{L}\rfloor+1}))\|_2)\right\|_2\right]\\
& \leq \mathbb{E}\left[ \left(\|\nabla \mathcal{L}(\theta^{t+1}(\Lambda^{\lfloor\frac{t}{L}\rfloor+2}))\|_2 + \|\nabla \mathcal{L}(\theta^{t}(\Lambda^{\lfloor\frac{t}{L}\rfloor+1}))\|_2\right) \left\| ( \|\nabla \mathcal{L}(\theta^{t+1}(\Lambda^{\lfloor\frac{t}{L}\rfloor+2}))\|_2 - \|\nabla \mathcal{L}(\theta^{t}(\Lambda^{\lfloor\frac{t}{L}\rfloor+1}))\|_2)\right\|_2\right]\\
&\leq 2 \mu\sigma \mathbb{E}\left[ \|(\theta^{(t+1)},\Lambda^{(t+2)}) -   (\theta^{(t)},\Lambda^{(t+1)})\|_2\right]\\
&\leq 2L\sigma \eta\eta_{\Lambda} \mathbb{E}\left[\left\|\left( \nabla \mathcal{L}(\theta^{t})+\psi^{(t)},  \nabla L_{CE}(\hat{\theta^{t+1}})+\xi^{(t+1)}   \right)\right\|_2\right]\\
& \leq 2L\sigma \eta\eta_{\Lambda} \mathbb{E}\left[ \sqrt{\| \nabla \mathcal{L}(\theta^{t})+\psi^{(t)} \|_2^2} +\sqrt{\| \nabla L_{CE}(\hat{\theta^{t+1}})+\xi^{(t+1)} \|_2^2}   \right]\\
&\leq  2L\sigma \eta\eta_{\Lambda} \sqrt{\mathbb{E}\left[ \| \nabla \mathcal{L}(\theta^{t})+\psi^{(t)} \|_2^2  \right]+\mathbb{E}\left[ \| \nabla L_{CE}(\hat{\theta^{t+1}})+\xi^{(t+1)} \|_2^2\right] }\\
&\leq 2L\sigma \eta\eta_{\Lambda} \sqrt{  \mathbb{E} \left[\| \nabla \mathcal{L}(\theta^{t})\|_2^2 \right]  + \mathbb{E}\left[
	\|\psi^{(t)}\|_2^2\right
	]    +\mathbb{E} \left[ \| \xi^{(t+1)}\|_2^2\right] +\mathbb{E} \left[ \| \nabla L_{CE}(\hat{\theta^{t+1}})\|_2^2  \right]}\\
&\leq 2L\sigma \eta\eta_{\Lambda} \sqrt{2\delta^2+2\sigma^2}\\
&\leq 2\sqrt{2(\delta^2+\sigma^2)}L\sigma\eta_{\Lambda} \eta.
\end{split}
\end{align}

According to the above inequality, we can conclude that our algorithm can achieve
\begin{align}
\lim_{t\rightarrow \infty} \mathbb{E}[\| \nabla \mathcal{L}(\theta^{t}(\Lambda^{\lfloor\frac{t}{L}\rfloor+1}))  \|_2^2]=0.
\end{align}
The proof is completed.

\end{proof}

\section{Additional details for Knowledge Distillation Experiments}

\subsection{Dataset Details}

Table~\ref{tab:textdatasplits} provides details of the various real-world datasets used in our experiments, and the partitioning of these datasets into train, validation (needed in our meta-learning procedure), and test data subsets. Wherever available, the existing splits provided by the source data were used; in other cases, 10\% of the training data was partitioned off for use as validation data.

\begin{table}[h]
    \centering
    \begin{tabular}{|l|l|l|l|l|l|}
    \hline
        Dataset & \#Classes & \#Instances & \#Train & \#Validation & \#Test \\ \hline
        CIFAR100 & 100 & 60000 &  45000 & 5000  & 10000   \\ \hline

         Stanford Cars & 196 & 16185 & 7330  &  814 &8,041   \\ \hline
     FGVC-Aircraft & 102 & 10200  &  6120 & 680  & 3400  \\ 
        \hline
    \end{tabular}
    \caption{Number of classes, Number of instances in Train, Validation and Test splits in the different  datasets} 
    \label{tab:textdatasplits}
\end{table}

\subsection{Additional Experimental Setup}

We ran experiments using an SGD optimizer with an initial learning rate of 0.05, the momentum of 0.9, and a weight decay of 5e-4. We divided the learning rate by 0.1 on epochs 150, 180 and 210 and trained for a total of 240 epochs. In all our knowledge distillation experiments we use temperature $\tau = 4$ and $\lambda_a =0.9$ (weights associated with KD loss) except in case of \model. We update the $\lambda$s every 10 epochs, therefore $L=10$. We ran all experiments on a single A100 GPU.  

\section{Additional experiments}\label{app:experiments}

\subsection{Distilling with multiple teachers} \label{sec:multi}
Although, the Knowledge distillation (KD~\cite{hinton2015distilling}) was introduced with only one teacher model and its corresponding loss, learning from multiple teachers has been shown to be useful \cite{mirzadeh2019improved,son2021densely,cho2019efficacy}. Specifically we adapt our method to improve DGKD~\cite{son2021densely}, where we perform knowledge distillation with multiple teachers and perform knowledge distillation with early stopped teachers \cite{cho2019efficacy}. Here, we redefine the Eq \ref{eq:distil} as,
\begin{align}
      \mathcal{L}_{student}(\theta,\mathbf{\Lambda}) = &  \sum_i \lambda_{p_i}\mathcal{L}_s\left(y_i, \xb_i|\theta\right) 
  + \sum_{k=1}^K \lambda_{a{k_i}} \nonumber \mathcal{L}_{KD}\left(y_i^{(S)}, y_i^{(T^k)}\right)
    \label{eq:multi_teach}
\end{align}
 We can also adapt it to knowledge distillation from an early stopped teacher as presented in~\cite{cho2019efficacy}, by  simply gathering multiple teacher checkpoints, and using them together as multiple teachers in knowledge distillation. We discuss the results and implication of this approach in Section~\ref{sec:multi_teacher}.
 
 \subsubsection{Knowledge Distillation with multiple teachers}\label{sec:multi_teacher}

In the multi teacher setup we use WRN-16-8 as a teacher model and perform knowledge distillation on DenseNet-40-12~\cite{DBLP:journals/corr/HuangLW16a} and WRN-16-1. Here we consider two settings to address the teacher student gap viz. performing knowledge distillation with a \textit{teacher stopped at some intermediate stage} \cite{cho2019efficacy} and learning from multiple teachers of different learning capacities (DGKD) \cite{son2021densely}.

The former approach raises a new challenge: how do we find an appropriate stopping point for the teacher, without having to train a large number of student models corresponding to teacher stopping points? We adapt our multi-teacher setting (Section~\ref{sec:multi}) to solve this as follows: we train a single student model, with \textit{multiple teacher models}, each stopped at intermediate points of the teacher training process. For this experiment, we trained a teacher model (WRN-16-8) on CIFAR-100 at the $80^{th},160^{th},200^{th}$ epochs as well as the final model, and trained two different student architectures using multi-teacher \model. Table~\ref{tab:early} shows that \model\ with early-stopped teachers consistently outperforms the standard KD as presented in \cite{cho2019efficacy} with one step knowledge distillation. 

DGKD introduced a stochastic variant where only a subset of teachers is introduced at each training step, determined by a binomial (hyperparamter) variable. This presents a need to control the contribution of each of the teachers in a systematic manner; therefore,  we adapt our multi-teacher setting (Section~\ref{sec:multi}) with WRN-16-3 as the additional teacher model. We present results in  Table~\ref{tab:early}, and observe that \model\ yields performance gains over simple DGKD.

\begin{table}[t]
\centering
    \begin{tabular}{c| c |c } \hline \hline
    Student Model $\rightarrow{}$ & WRN-16-1 & DenseNet-40-12 \\ \hline
    Method $\downarrow$ &   & \\ \hline
{KD} & 66.47 & 76.57 \\ \hline

 {\model\ with early} &  \multirow{2}{2em}{67.69}  &  \multirow{2}{2em}{76.79} \\
 {stopped teachers}&&\\\hline
 {DGKD} & 67.88 & 76.86 \\ \hline
  {DGKD+ \model\ } & 68.58 & 77.66\\ \hline
 \hline 
    \end{tabular}
    \caption{Learning from early-stopped teacher: \model\ was trained with a range of teacher checkpoints (early stopping) in the multi-teacher setting. $\lambda$s obtained from \model's meta learning process helps  identify a good teacher. Similarly, \model\  beats SOTA DGKD while learning from different teacher model. Here, the teacher model was WRN-16-8, and CIFAR-100 was used as dataset}
    \label{tab:early}
    \end{table}

\subsection{$\lambda$s for better generalisation} 

We now explore the \textit{self-distillation} setting~\cite{furlanello2018born,DBLP:journals/corr/abs-1908-01851}, where the teacher and student models have identical architectures, and the goal is to train a student with \textit{better generalization accuracy} than the teacher, through the use of the distillation loss for regularization. In Table~\ref{tab:self}, we present the results of self-distillation experiment on the two datasets {\em viz.}, CIFAR100 and the FGVC-Aircraft datasets on a WRN-16-8 model.

The first row presents training results based on the standard cross-entropy loss and the second row presents the results in standard knowledge distillation setup. Third row of
table presents results with  Platt-scaling \cite{pmlr-v70-guo17a}, which rescales test outputs based the validation set. Here we apply Platt-scaling over the knowledge distillation setup presented in the second row. The final row shows that adaptive mixing via \model\ further improves upon self-distillation, thereby making it a potentially valuable tool in the traditional supervised learning setting in addition to teacher-student transfer for training smaller, more efficient student models. Poor platt scaling results indicate that use validation data doesn't always strengthen the baselines. 

\begin{table}[H]
    \centering
    \begin{tabular}{c| c c } \hline \hline
    Dataset $\rightarrow{}$ & CIFAR100 & FGVC-Aircraft   \\ \hline
    Method $\downarrow$ &   & \\ \hline
{CE loss alone} & 77.52 & 63.75 \\ \hline
 {Self-distillation}& 79.12 & 66.93  \\ \hline 
 {Self-distillation}&  \multirow{2}{2em}{79.01}& 
 \multirow{2}{2em}{66.12}   \\ 
 {+ Platt-Scaling}&  &  \\ \hline 
 {Self-distillation}&  \multirow{2}{2em}{79.41}& 
 \multirow{2}{2em}{67.44}   \\ 
 { + \model\ } & & \\\hline 
 \hline 
    \end{tabular}
    \caption{Self-distillation: Compared to traditional supervised learning (CE loss alone) self-distillation show higher accuracy, as reported by earlier work. We show that \model\ is easily applicable to the self-distillation setting, and provides additional gains in performance as oppose to other ways of using validation set such as Platt-scaling}
    \label{tab:self}
    
\end{table}

\subsection{Ablation study with temperature $\tau$ and Frequency of $\lambda$ updates $L$}

\begin{figure}[H]
\centering
\includegraphics[width = 0.20\textwidth]{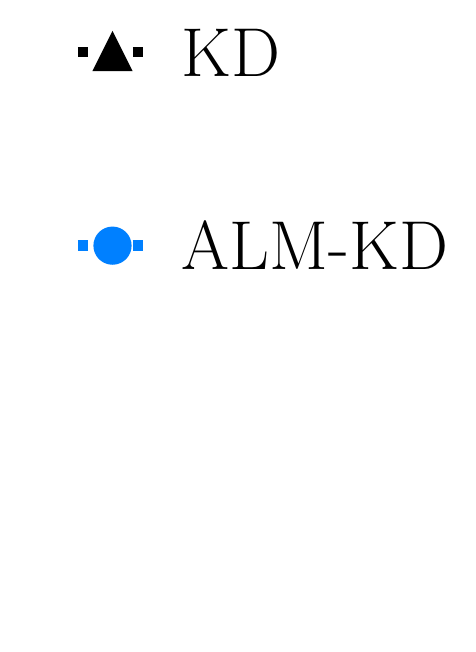}
\centering
\begin{subfigure}[b]{0.4\textwidth}
\centering
\includegraphics[width=0.87\textwidth]{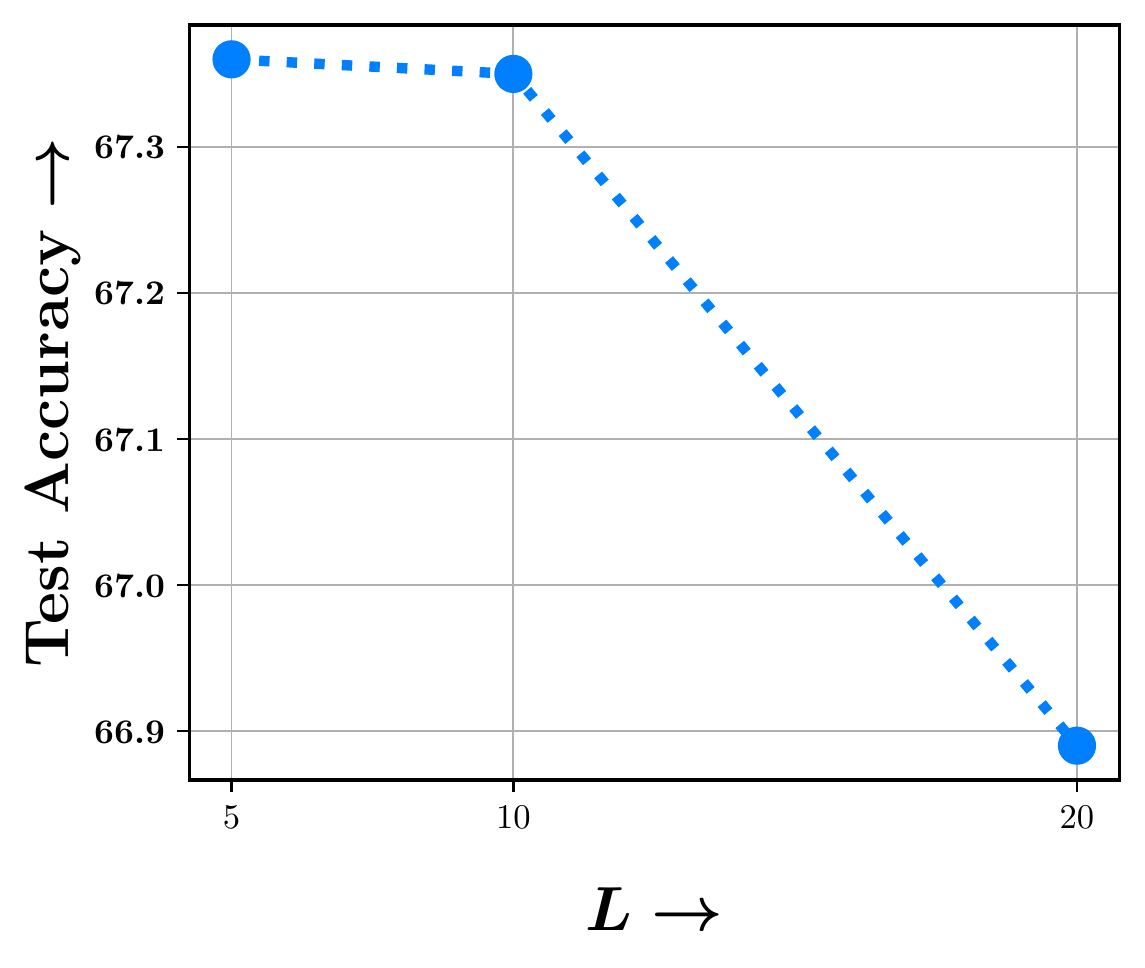}
\caption{\centering Test accuracies achieved by  \newline \model\ verse $L$}
\label{fig:fre_L}

\end{subfigure}
\begin{subfigure}[b]{0.35\textwidth}
\centering
\includegraphics[width=\textwidth]{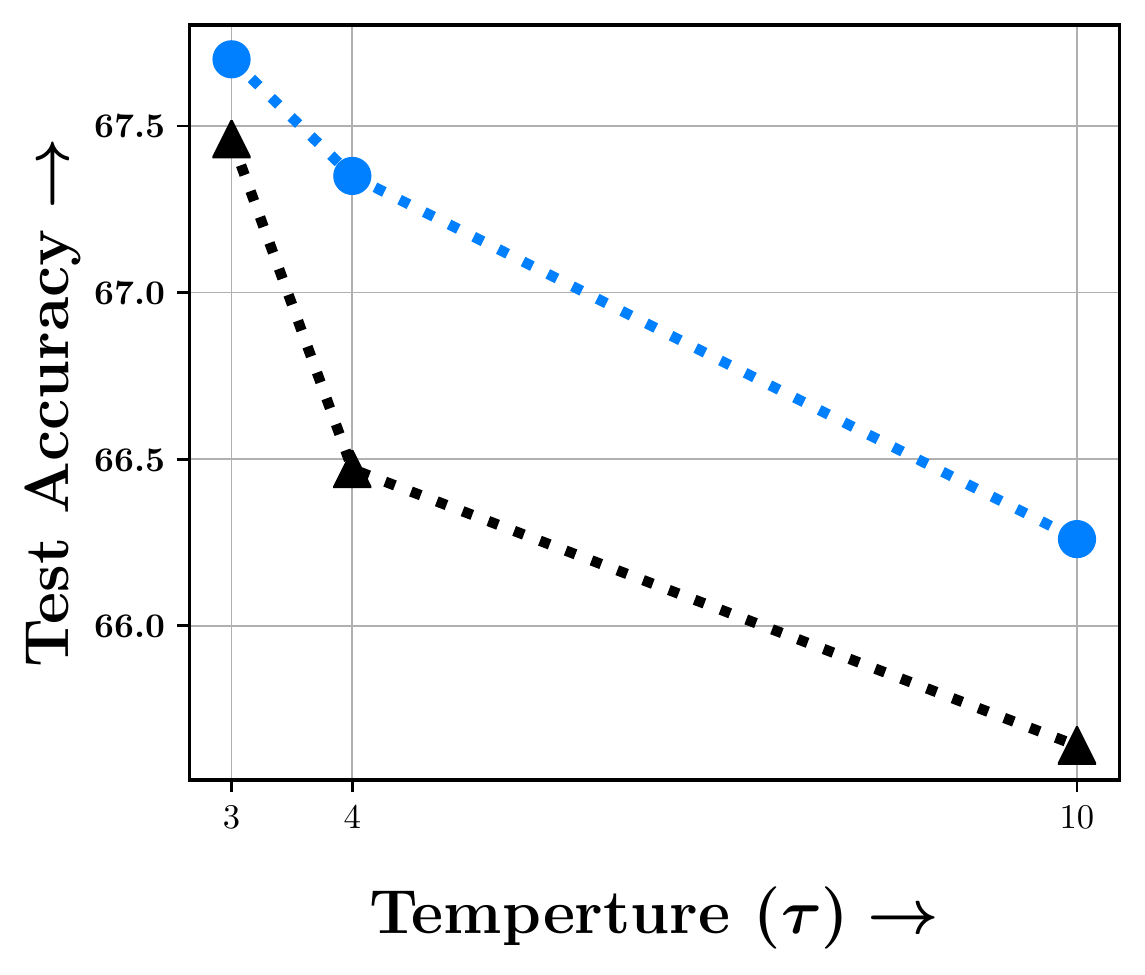}
\caption{\centering Test accuracies achieved by \newline \model\  verse temperature $\tau$ }
\label{fig:temperature}
\end{subfigure}
\caption{Effect on test accuracies achieved by \model\ as we vary temperature $\tau$ and $L$ while performing knowledge distillation on CIFAR100 dataset with WRN-16-8 as the teacher and WRN-16-1 as the student model.}
\label{fig:abliation}
\end{figure}

Figure \ref{fig:fre_L} shows change in test accuracies achieved by \model\ as we vary $L$ parameter which controls how often lambdas are updated. Here we present results for $L = 5,10,20$. Although, there isn't significant drop in test accuracy when we update $\lambda$s every 10 epochs instead of 5, there is a slight drop when we update $\lambda$s every 20 epoch. $L$ controls the trade-off between the time spent on updating lambdas verse improving the model performance. Here we find the $L=10$ is best as it saves time by not updating $\lambda$s too often while also achieving comparable performance as of $L=5$.   
Figure \ref{fig:temperature} we study the effect on the test accuracies achieved by \model\ as we vary temperature $\tau$ parameter used to control the softening of the KD loss. Clearly, across different $\tau$ values \model{} outperforms the standard KD and therefore \model's performance gains are not effected by change in $\tau$.

\subsection{Additional analysis on $\lambda$ values learnt in nosiy setting}\label{lam_analysis}

Similar to the distribution presented in Figure \ref{distill_diff}, we present sum weights associated with the distillation loss and the supervision loss ($\lambda_a$ and $\lambda_p$ respectively), when \model{} is used to perfrom KD in presence 40\% label noise injected into the CIFAR100 dataset.  Here, too, we use ResNet110 as teacher and ResNet8 as student. Here we see \model{} assigns more weight to the cleaner points which helps us understand \model's superior performance in presence of noise in Figure \ref{fig:noise_distill}. We also see a small percentage of noisy points getting higher weights. This could be because of presence of learning opportunity from teacher's output.

\begin{figure}[H]
\centering
\includegraphics[width =\linewidth ]{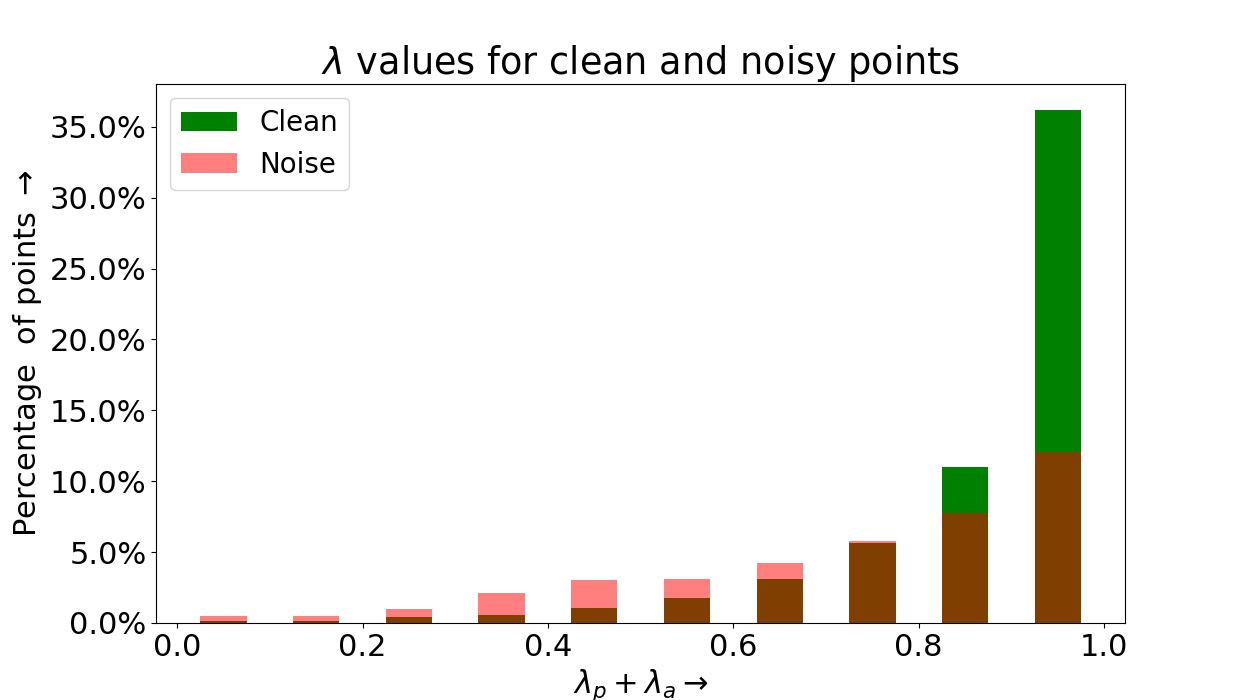}
\caption{Distribution of sum of  $\lambda_a$ weight associated with the knowledge distillation loss  and $\lambda_p$ weight associated with the cross entropy loss, obtained using \model{} while performing knowledge distillation with Resnet8 as the student model and Resnet110 the teacher model on CIFAR100 dataset with 40\% label noise. }
\label{distill_plus}
\end{figure}

\subsection{How does \model\ work in KD?}

\begin{figure}[H]
\begin{center}
\includegraphics[width=0.49\textwidth]{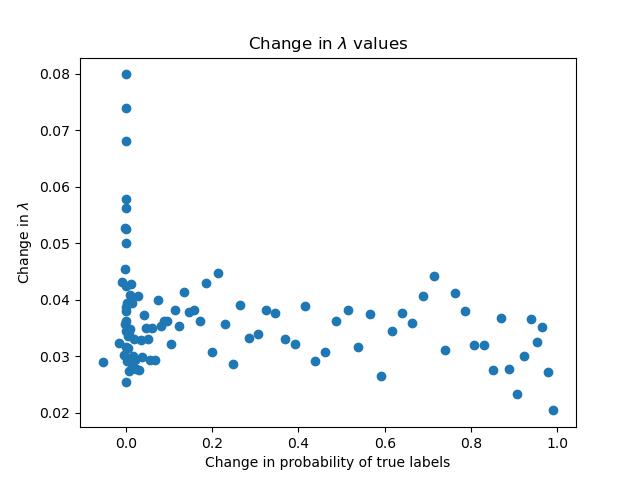}
\end{center}
\caption{Difference in $\lambda$s obtained for early stopped teacher {\em vs.} final teacher model for the corresponding change in true label confidence on  CIFAR-100 dataset while training the WRN-16-1 Student model with the WRN-16-8 teacher model.}

\label{fig:early_vs_final}
\end{figure}


We also took a closer look at the above hypothesis: that early stopping of the teacher helps by creating `simpler', easy-to-mimic teacher outputs. We used two WRN-16-8 teachers--early stopped, and final, on CIFAR-100 -- to train WRN-16-1 Student models corresponding to each teacher. We then examined the \textit{change in lambda value} ({\em i.e.}, teacher weight) across the two student models, as a function of \textit{change in teacher confidence}, {\em i.e.}, the probability output by the teacher to the ground-truth label in training data. Figure~\ref{fig:early_vs_final} shows a very interesting result: as the teacher increased in confidence, the $\lambda$ values chosen by \model\ \textit{decreased} whereas if teacher confidence was largely unchanged, the $\lambda$ values were increased. This suggests that the primary challenge in the final teacher model is \textit{overfitting}, with even noisy or ambiguous labels being confidently predicted by the teacher. Our validation-based objective, however, is able to identify those instances that the teacher is overconfident on (as they do not improve validation accuracy), and able to downweight the teacher in those instances.

\section{Additional details for Limited supervision and rule-denoising Experiments}
Here, we provide additional details about the datasets used in rule-denoising experiments. We used three dataset in our experiments, namely, YouTube, SMS and IMDB. In addition to the features, we have access to $m$ rules of labelling functions (LFs). In Table \ref{tab:stats}, we provide statistics of these LFs as well as size of labeled and unlabeled set. We borrow LFs from the SPEAR and our experimental setting such as batch size, learning rates are same as SPEAR \cite{maheshwari-etal-2021-semi} to ensure fair comparison.

\begin{table}[th!]
\centering
\begin{adjustbox}{width=0.9\textwidth}
\begin{tabular}{lccccccc} 
\toprule
Dataset & $|\mathcal{L}|$ & $|\mathcal{U}|$ & \#Rules/LFs & Precision & \%Cover  & \%Conflicts &  \textbar{}Test\textbar{} \\ 
\toprule
YouTube & 100 & 1586 & 10 & 75 & 86.6& 30.1 & 250 \\
SMS & 69 & 4502 & 73 & 97.3 & 39.3 & 0.67 & 500 \\
IMDB & 284 & 852 & 25 & 80 & 48.6 &  11.1 &500 \\
\bottomrule
\end{tabular}
\end{adjustbox}
\caption{Statistics of datasets and their rules/LFs. Precision refers to micro precision of rules. \%Cover is the fraction of instances in $\mathcal{U}$ covered by at least one LF. \%Conflict denotes the fraction of instance covered by conflicting rules among all instances. Size of Validation set is equal to $|\mathcal{L}|$.}
\label{tab:stats}
\end{table}

\section{Rule-denoising objective}\label{rule-objective}
Here, we describe the individual loss components borrowed from SPEAR \cite{maheshwari-etal-2021-semi}. Further, we define our adaptive mixing loss which forms our overall objective function.

\noindent \textbf{First Component (L1): } Standard cross-entropy loss on  $\Dcal$ for the model $P_\theta^f$ :  $L_{CE}\left(P_\theta^f(y|\bfx_i), y_i\right) = -\log\left(P_\theta^f(y=y_i|\bfx_i)\right)$

\noindent \textbf{Second Component (L2): } The second component $LL_s(\phi|\Dcal)$ is the (supervised) negative log likelihood loss on the labeled set $\Dcal$:
    $LL_s(\phi|\Dcal) = - \sum \limits _{i=1}^{N} \log P_\phi(\bfl_i, y_i)$\\

\noindent \textbf{Third Component (L3): } The third component $L_{CE}\left(P_\theta^f(y|\bfx_i), g(\bfl_i)\right)$ is the cross-entropy of the classification model using the hypothesised labels from CAGE ~\cite{cage} on $\Ucal'$. CAGE is a generative graphical model that assigns parameter $\phi_j$ for each rule and share it across $y$. (Please refer Appendix for the complete formulation.) Using the LF-based graphical model $P_\phi(\bfl_i, y)$ as: $g(\bfl_i) = \mbox{argmax}_y  P_\phi(\bfl_i, y)$\\

\noindent \textbf{Fourth Component (L4):} The fourth component $KL(P_\theta^f(y|\bfx_i), P_\phi(y|\bfl_i))$ 
is the Kullback-Leibler (KL) divergence between the predictions of both the models, {\em viz.}, feature-based model $f_\theta$ and the rule-based graphical model $P_\phi$  summed over every example $\bfx_i \in \Ucal' \cup \Dcal'$. We try and make the models agree in their predictions over the union of the labeled and unlabeled datasets.

\subsection{Adaptive Loss Mixing for limited supervision and rule-denoising}
In our joint objective, feature based classification model  $f_\theta(\bfx)$ while second component (L2) trains the rule-based model. L1 and L2 works on $\Dcal'$, L3 works on $\Ucal'$ and L4 component works on $\Dcal' \cup \Ucal'$. Therefore, joint objective can be rewritten using instance wise weights $\lambda_{p_i}$ and $\lambda_{a_{1,i}}$introduced in Eq. \ref{eq:instance} as, 
\begin{align}
    \Lcal = 
    \begin{cases}
     \lambda_{p_i}*L_{CE}\left(P_\theta^f(y|\bfx_i), y_i\right) + \lambda_{a_{1,i}}KL(P_\theta^f(y|\bfx_i), P_\phi(y|\bfl_i)) & \text{if}\ (\xb_i,y_i,l_i)  \in  \Dcal'\\ 
     \quad \quad+ LL_s(\phi|\xb_i,y_i), & \\
      \lambda_{p_i}*L_{CE}\left(P_\theta^f(y|\bfx_i), g(\bfl_i)\right) + \lambda_{a_{1,i}}KL(P_\theta^f(y|\bfx_i), P_\phi(y|\bfl_i)) &  \text{if}\ (\xb_i,y_i,1_i)  \in  \Ucal'\\ 
    \end{cases}
\end{align}

Thus our primary objective changes based on whether the point belongs to the labelled set $\Dcal'$ or the unlabelled set $\Ucal'$. 

\section{Synthetic experiments}

We explore the performance and characteristics of our approach in synthetic data settings, to derive insight into the mechanisms by which \model. 

\begin{figure}[!h]
\centering
\begin{subfigure}[b]{0.48\textwidth}
\centering
\includegraphics[width=6cm,height=4.4cm]{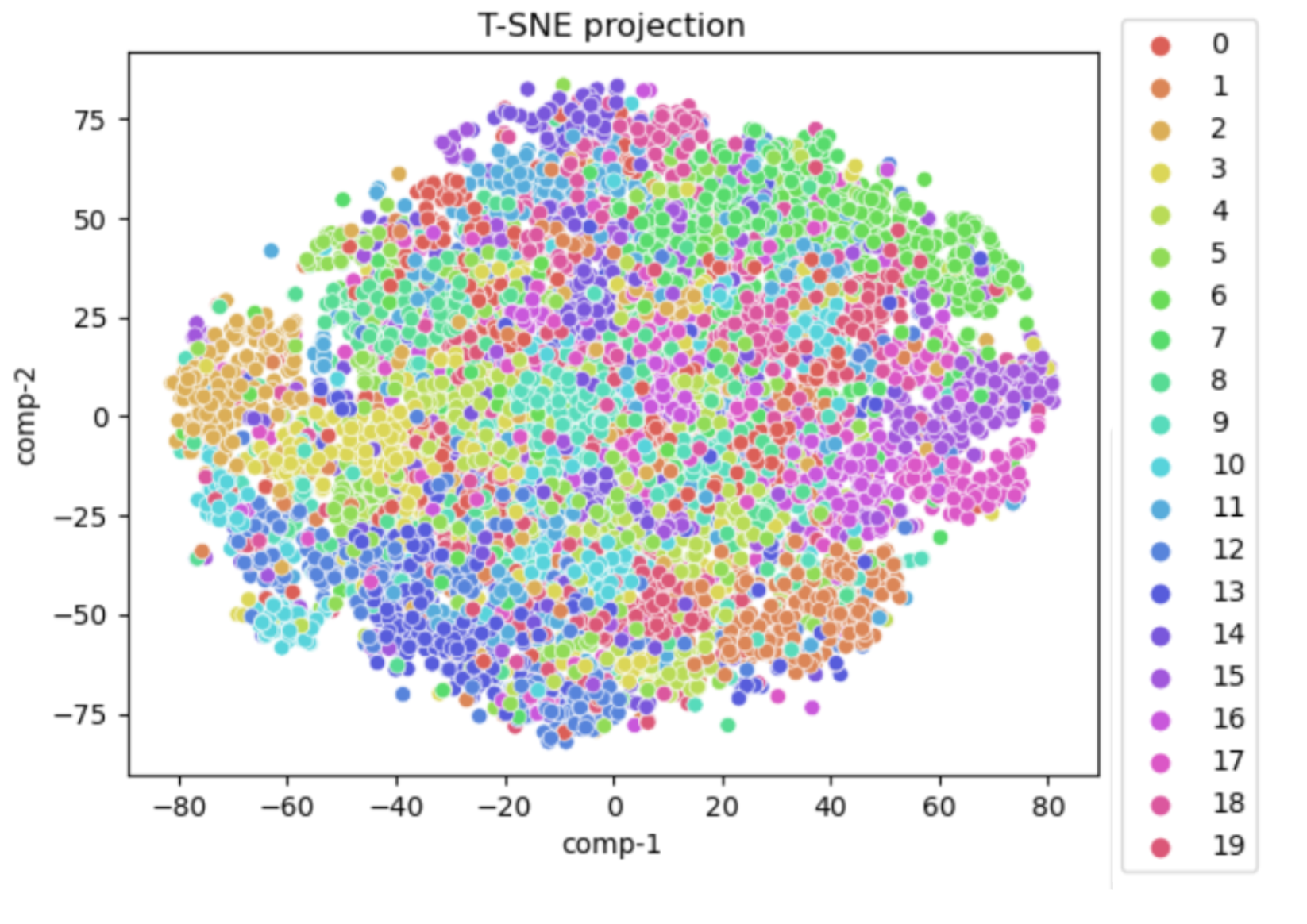}
\caption{t-SNE plot of the synthetic data used.}
\label{fig:synthdata}
\end{subfigure}
\hfill
\begin{subfigure}[b]{0.48\textwidth}
\centering
\includegraphics[width=6cm]{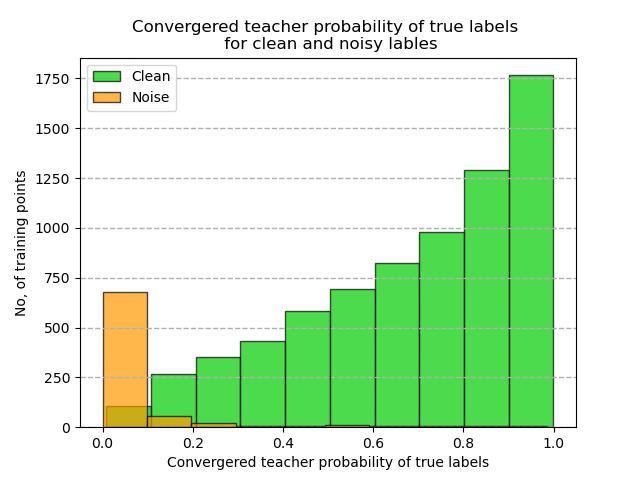}
\caption{\centering Teacher confidence on ground truth for noise vs clean points}
\label{fig:lambda_noise_full}
\end{subfigure}

\begin{subfigure}[b]{0.48\textwidth}
\centering
\includegraphics[width=6cm,height=4.4cm]{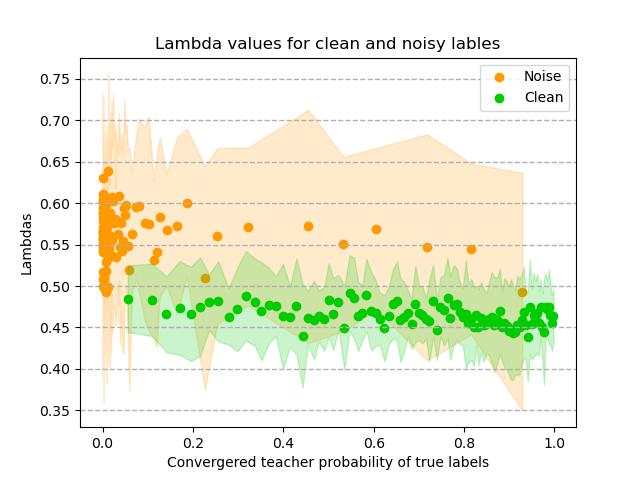}
 \caption{Relationship between teacher confidence, label noise, and learned $\lambda$s with mean and SEM error bars over 15 runs. As elaborated in Section~\ref{sec:counterNoise}, the teacher assigns relatively lower probabilities to noisy labels, and the meta-learning process in \model\ assigns higher $\lambda$ to noisy data points.}
    \label{fig:lambda_noise}
\end{subfigure}
\hfill
\begin{subfigure}[b]{0.48\textwidth}
\centering
\centering
    \includegraphics[height=2in]{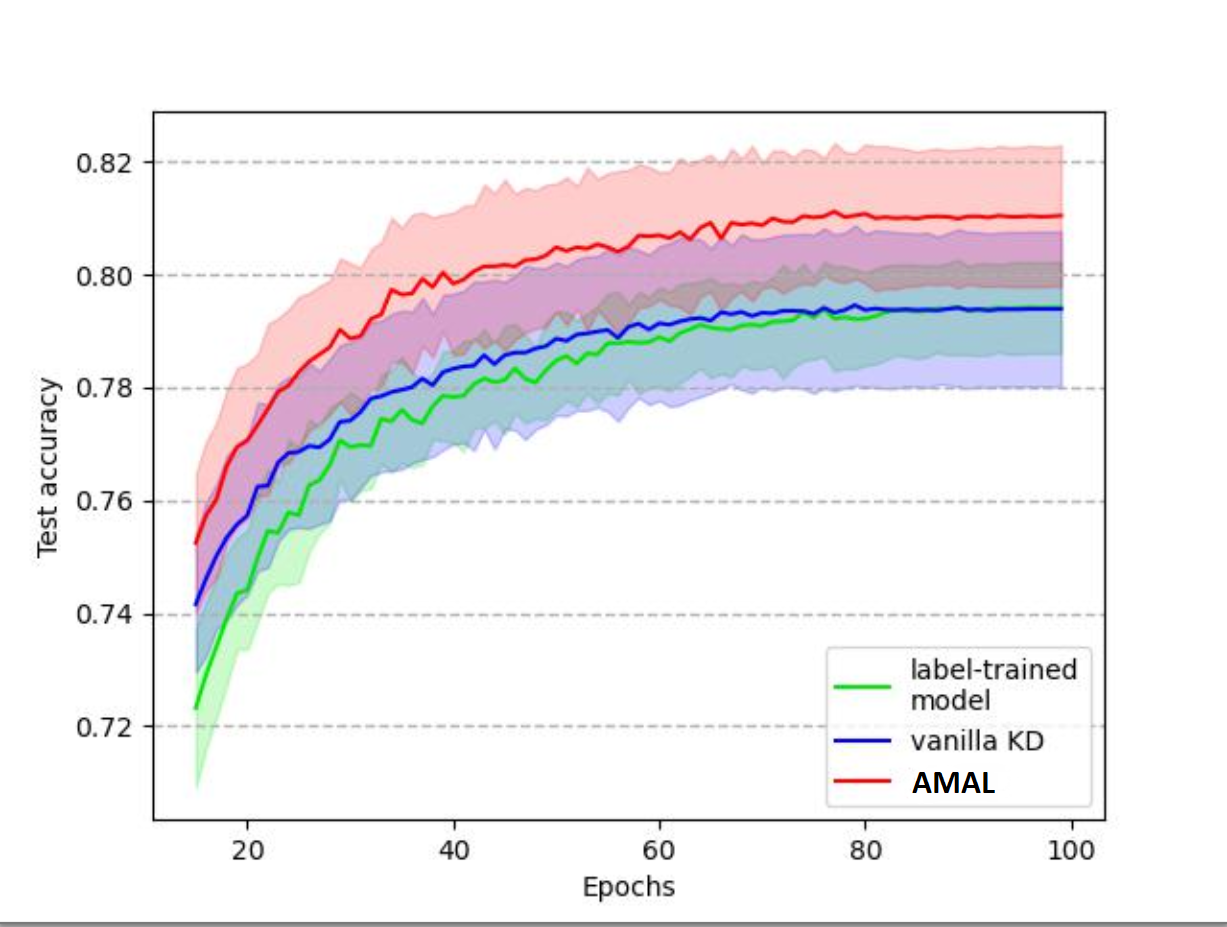}
    \caption{Learning trajectories: The figure shows generalization of test set accuracies (along with SEM bars) for vanilla KD, label-trained model, and \model, as a function of training epochs. All curves are averages over 50 runs.}
    \label{fig:generalization}
\end{subfigure}

\caption{Synthetic experiment details and analysis}

\label{fig:syn_experiments}
\end{figure}

\textbf{Synthetic data generation: }
We use the standard \textbf{sklearn.datasets} package~\cite{scikit-learn} to generate synthetic data with 14 features and 20 classes. The generated data has 8100 training points, 900 points in validation set and 1000 points in the test set. We randomly flip labels of 10\% of the training data points to introduce noise. In Figure~\ref{fig:synthdata}  shows a $t$-sne projection of a synthetically generated dataset (generation details in the main body of the paper); as can be seen, the 20 classes have some spatial cohesiveness but also significant overlap, making it a nontrivial learning task to classify instances into their respective labels.

\subsection{$\lambda$s can counter label noise} \label{sec:counterNoise}
Through this experimental setup, we examine the relationship between the teacher confidence, the label noise, and the learned $\lambda$ values of data points. We grouped training data points into buckets, based on the probability assigned by the teacher to its ground-truth label, and computed the average learned $\lambda$ per bucket; this averaging process was done separately for the instances with and without injected label noise. Finally, the experiment was repeated with 50 random seeds, and the averages as well as standard error of the mean (SEM) bars across those 50 runs are presented in Figure~\ref{fig:lambda_noise}. We see two emerging patterns: Firstly, that as expected, the teacher  assigns overall lower probabilities to the noisy labels in comparison to the other data points. Secondly, we see a trend wherein $\lambda$s learned for noisy labels are overall higher than for clean labels. This is because the meta-learning process learns that the ground-truth labels on those data points contribute poorly to generalization on the validation data, whereas, for such instances, the teacher probability is more informative than the ground-truth label. In Figure~\ref{fig:lambda_noise_full}  shows  the distribution of converged teacher probabilities for noisy and clean labels, showing, as expected, that the clean label data have a broad distribution of teacher probabilities for ground-truth label, with a right skew, whereas the noisy labels have a sharply leftward skew ({\em i.e.}, very low teacher probabilities for ground-truth labels). This nicely complements the data in the left panel, showing that for noisy labels, and in general for less-confident teacher signals, the learned $\lambda$ values are higher, indicating that the teacher has more informative content ({\em e.g.}, instance hardness, label ambiguity) than the ground-truth label in those scenarios.

In a second experiment, we further examined the contributions of the meta-learning procedure to test-set generalization. In Figure~\ref{fig:generalization}, we present the test set accuracy as well as standard error of the mean (SEM) bars for 
various models as a function of training data epoch. Each curve is the average of 50 random synthetic data simulation-based training runs. We see that compared to the student model trained on the ground-truth data ({\em i.e.} the {\color{green} label-trained model}), as well as the student trained with fixed $\lambda =0.9$  ({\em i.e.} {\color{blue} vanilla KD}), {\color{red} \model}\ 
learns faster, and converges to a higher test accuracy, driven by the adaptive loss mixing approach ({\em c.f.} Section~\ref{sec:adaptiveLossMixing}).

\end{document}